\pdfoutput=1

\PassOptionsToPackage{xcdraw,table}{xcolor}

\documentclass[11pt]{article}
\usepackage{comment}
\usepackage[final]{acl}
\usepackage{times}
\usepackage{latexsym}

\usepackage[T1]{fontenc}

\usepackage[utf8]{inputenc}
\usepackage{booktabs}
\usepackage{microtype}

\usepackage{inconsolata}

\usepackage{graphicx}
\usepackage{float}
\usepackage{color}
\usepackage{colortbl}
\usepackage{xspace}
\newcommand{\lcfo}{\textsc{LCFO}\xspace}

\usepackage{tabularx}%
\usepackage{siunitx}
\usepackage{multirow}
\usepackage{arydshln}

\usepackage{colortbl}
\colorlet{tableheadcolor}{gray}
\colorlet{tablerowcolor}{gray!10}
\newcommand{\rowcol}{\rowcolor{tablerowcolor}} %
\title{{\lcfo:}\\[7pt]\Large Long Context and Long Form Output Dataset and Benchmarking}

\author{Marta R. Costa-juss\`a, Pierre Andrews, Mariano Coria Meglioli, Joy Chen,\\
\textbf{Joe Chuang, David Dale, Christophe Ropers, Alexandre Mourachko,}\\ \textbf{Eduardo Sánchez, Holger Schwenk, Tuan Tran, Arina Turkatenko,  Carleigh Wood}\\
FAIR, Meta\\
\tt{\{costajussa,mortimer,mfcoria,joyqchen}\\\tt{joe.chuang,daviddale,chrisropers,alexmourachko}\\\tt{eduardosanchez,schwenk,tuantran,arinatur,carleighwood\}}@meta.com}

\usepackage{algorithm}
\usepackage[utf8]{inputenc}

\usepackage{enumitem}

\usepackage{tablefootnote}

\newcommand{\llama}{\textsc{LLaMA}\xspace}

\begin{document}
\maketitle

\begin{abstract}

This paper presents the Long Context and Form Output (LCFO) benchmark, a novel evaluation framework for assessing gradual summarization and summary expansion capabilities across diverse domains. \lcfo{} consists of long input documents (5k words average length), each of which comes with three summaries of different lengths (20\%, 10\%, and 5\% of the input text), as well as approximately 15 questions and answers (QA) related to the input content. Notably, \lcfo{} also provides alignments between specific QA pairs and corresponding summaries in 7 domains.

The primary motivation behind providing summaries of different lengths is to establish a controllable framework for generating long texts from shorter inputs, i.e. summary expansion. 
To establish an evaluation metric framework for summarization and summary expansion, we provide human evaluation scores for human-generated outputs, as well as results from various state-of-the-art large language models (LLMs). 

GPT-4o-mini achieves best human scores among automatic systems in both summarization and summary expansion tasks ($\approx$ +10\% and +20\%, respectively). It even surpasses human output quality in the case of short summaries ($\approx$ +7\%).
Overall automatic metrics achieve low correlations with human evaluation scores ($\approx$ 0.4) but moderate correlation on specific evaluation aspects such as fluency and attribution ($\approx$ 0.6).


\end{abstract}

\section{Introduction}

Robust long text generation capabilities are required to meet user demand for extensive content creation, including story writing and essay composition \citep{xie-riedl-2024-creating}, which is why recent models such as GPT-4 \citep{openai2024gpt4technicalreport} are expanding the output lengths from 4k tokens in GPT-4o to 64k in the latest versions. 

\begin{figure*}[!ht]
\centering
{\includegraphics[width=\textwidth]{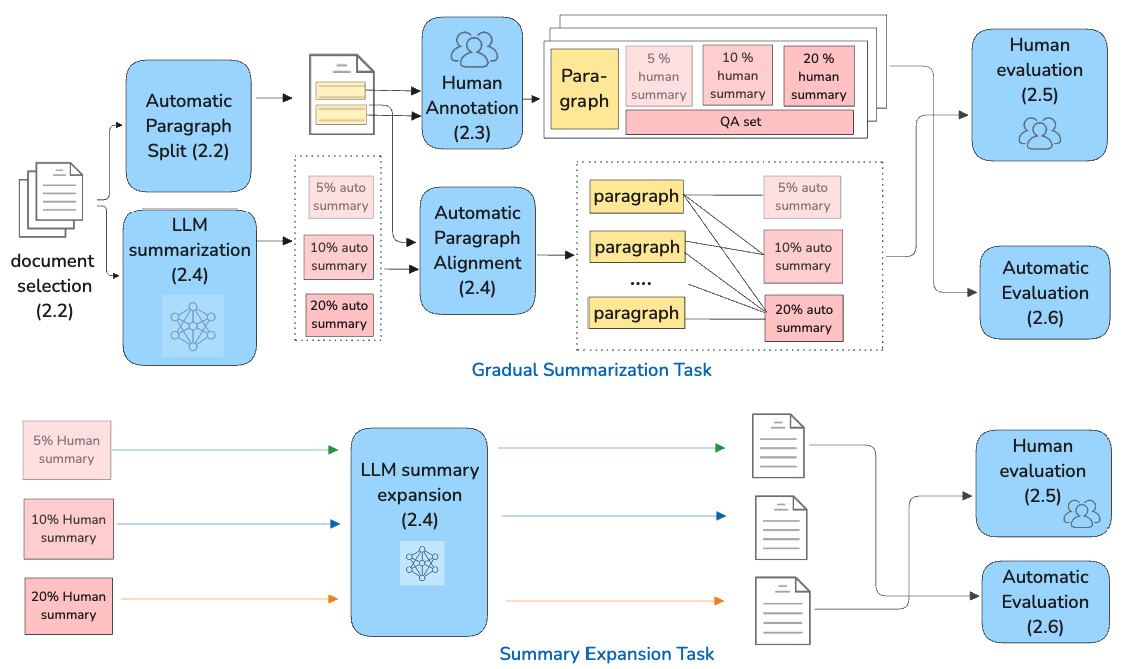}}
\caption{ LCFO annotation and evaluation framework for the gradual summarization (top) and summary expansion (bottom) tasks.  The resulting benchmark can also be used for other tasks such as  reading comprehension and automatic evaluation metric development. In gradual summarization, long documents are split into paragraphs and annotated by humans to generate 3 level of summaries, plus a questions-answers set. Additionally, different LLMs were used to construct automated summaries at corresponding levels. In summary expansion, different human summaries are prompted to LLMs to generate the output targeting the same length as the source document. The components are described in different sections (marked in parentheses) }
\label{fig:schema}
\end{figure*}

However, evaluating the performance of Large Language Models (LLMs) on summarization and summary expansion tasks (see definitions in Section \ref{sec:definitions}) is particularly challenging, especially when it comes to summarizing very long input documents and generating either long summaries or long summary expansions. Although there is a lot of work and interest in studying summarization evaluation (e.g. \citealp{zhang2024systematicsurveytextsummarization}), evaluation of long text outputs is an emerging area \citep{que2024hellobenchevaluatinglongtext}. Indeed, long-context input processing tasks (such as summarization or comprehension question answering applied to long documents) and long-form output production both involve high cognitive loads for humans. This may be why evaluation work in these areas is not as mature as in others. 


To complement the summarization task, we include an ``inverted'' task of \textit{summary expansion}: generating a longform text based on its shorter sketch. It requires more creativity than summarization (which can be approached extractively) but imposes more constraints than open-ended language modeling. A practical application for this task could be with a writing assistant expanding the text sketch created by the human writer into a longer output that the human subsequently revises. Another potential use is “back-translation” for summarization, with a model training to summarize the synthetic expanded documents into their original short forms.



The \lcfo{} dataset that we are presenting in this paper is an exclusively human annotated benchmark and challenge dataset involving natural language understanding and generation across multiple domains in the aforementioned tasks. Our dataset is carefully manually crafted with human revision from the selection of the documents to the final annotation, without relying on LLMs at any point.  We provide detailed linguistic guidelines and abstractive QA. Furthermore, we provide another set of linguistic guidelines to evaluate the tasks of summarization and summary expansion.

Together, we present the \lcfo{} benchmark. Given a source of long structured documents, we generate multiple long outputs and associated QA pairs, and we evaluate human and model outputs with human annotations. 
The main contributions of this work are the following (see Figure \ref{fig:schema} for a schematic representation of the dataset and tasks):

\begin{itemize}[noitemsep]
    \item Dataset creation with structured inputs and alternative references; each input document is associated with 3 summaries of different lengths (20\%, 10\%, 5\%). Gradual summarization is useful in that it provides both long and short summaries references. Moreover, the availability of summaries of various length
is expected to improve the development of summary expansion approaches, which allows us to provide a more controllable summary expansion framework.
\item A set of QA pairs for each input document aligned with each of the different length summaries. Our QA pairs are in free-form short-answer format (i.e. not multiple choice) and are of the abstractive type (i.e. they are not copies of parts of the source document). This QA can potentially be used to evaluate the model outputs, based on the appropriateness of the responses, similarly to previous proposals \citep{wang-etal-2022-squality}. 
\item Selection of automatic metrics. 
 Most of these metrics can be used to evaluate at the paragraph level, and more widely can be used to evaluate summary expansion or long-form generation in multiple tasks and languages.  
\item Evaluation of several LLMs on our dataset both automatically and manually; and evaluation of automatic metrics on summarization and summary expansion.

\end{itemize}


\section{The LCFO Creation Framework: Principles and Methods}

This section describes the detailed principles and methods that underlie LCFO (Figure \ref{fig:schema} (right)).

\subsection{Definitions}
\label{sec:definitions}

\paragraph{Long context/form.} We define long text as text that exceeds 5k words. For reference, the attention span for reading and taking notes has been considered 10 to 15 minutes since a 1978 seminal paper \citep{hartley-davies-1978}. More recent research cited in a survey paper \citep{Bradbury-2016} shows that attention paid to lectures declines significantly after 20 minutes. The tasks we describe here are closer to note taking than lecture attendance; therefore, we should keep the 10--15 minute reference. If we base ourselves on the reported reading average time for first-language, secondary-education level readers (125 words per minute on average), we can consider that we will reach high cognitive load at around 1.5k words, and peak cognitive load at around 2.5k words. The quality of cognitive processing then starts to decrease significantly after.

\paragraph{Structured/hierarchical.} The input and output documents are partitioned into sections and, if necessary, nested sub-sections. Their structure is determined by task- and domain-specific guidelines. 

\paragraph{Gradual summarization (GS).} The input is a long document (defined here as 5k words or more). The summarization task can be described as the act of generating a much shorter corresponding document that includes the essential information contained in the long document and the same logical structure linking the various pieces of essential information. The summarization task presented here consists in taking long documents as inputs and generating three corresponding summaries of length that represent 20\%, 10\%, or 5\% of the input document.\footnote{We did not explicitly instruct the human annotators to generate the shorter summaries hierarchically, by compressing the longer ones, but this is what they apparently did most of the time. Hence the term ``gradual'': our data allows gradually interpolating between various levels of details in a text.}

\paragraph{Summary expansion (SE).} The input is a short and concise document that has similar properties to those of a summary (that is, it is mainly a standalone document that abstracts from details). The summary expansion task can be described as the act of generating a much longer document that preserves the essential elements found in the corresponding short document as well as the logical structure that connects such elements. More specifically, the task presented here consists in taking summaries as inputs and generating 3 long documents of different lengths. Each of the 3 lengths is set such that an input summary represents either 5\%, 10\%, or 20\% of its respective expanded documents. As this is a more freely generative task, an additional requirement to be taken into consideration is that of coherence (for example, the detailed information included in one generated sentence should not contradict that included in another sentence). We prompt the model only with one single summary and specify the length of the output we want (5x, 10x...). We do not include in the prompt longer summaries or the whole document, since they may be a possible output.

\subsection{Data selection and preprocessing}

\paragraph{Selection.} We select input documents that cover different domains, which meet the desired average length and the structure requirement.  
\begin{itemize}[noitemsep, nolistsep]
    \item We cover 7 domains including politics, news, Wikipedia, scientific, literature, conversational, and legal documents, with the format of documents and conversations. We source from 10 datasets:
LexGLUE \citep{chalkidis-etal-2022-lexglue}, BookSum \citep{kryscinski-etal-2022-booksum}, SQuality \citep{wang-etal-2022-squality}, FacetSum \citep{meng-etal-2021-bringing} JRC-Acquis \citep{steinberger-etal-2006-jrc}, MultiUN, Wikipedia, GovReport \citep{huang-etal-2021-efficient}, Summscreen \citep{chen-etal-2022-summscreen}, and Seahorse \citep{clark-etal-2023-seahorse}.  The correspondence between the source data sets and the domains is shown in Table \ref{tab:dataset-fields}.

\item The source documents are selected to be on average 5k words / document. We prefer documents with a hierarchical structure and containing relatively few numbers\footnote{We de-prioritize documents with many numbers (such as Wikipedia pages with large tables of various statistics), because our focus is less on structured data and more on natural language, and also because comparing large sets of numbers in the source and in the summary would be a difficult task for human annotators. }, which are better suited for summarization and summary expansion tasks. 
\item We prioritize recent documents when the domains allow (e.g., Wikipedia, where articles have a significant amount of new information since 2024). We preprocess documents in structured domains to provide a flattened structure while keeping hierarchical markers that are readable to annotators and models. It also ensures a consistent format across datasets. 
\item We filter out documents that contain toxicity using the ETOX package \citep{costa-jussa-etal-2023-toxicity} and add manual verification to ensure the high integrity of the selected documents.
\end{itemize}
\paragraph{Preprocessing.} 
To reduce the cognitive load for human annotation, we split paragraphs automatically (APS). Details on this paragraph splitting differ from corpus to corpus and are reported in the Appendix \ref{app:paragraph}.

\subsection{Human summary and QA-pair generation }

To obtain human-written summaries of long-form texts,  detailed guidelines are developed (Appendix \ref{human:summguide}). All summary writers must be native English speakers and have writing or editing experience.

These writers receive 252 long form documents (each around 5k words), and they are asked to read each document in its entirety and write three summaries for each document: the first summary representing around 20\% of the length of the source document; the second and the third summaries representing further summarization --- around 10\% and 5\% of the source document, respectively. When writing these summaries, the writers are tasked with compressing and retaining all the core ideas of the source. Giving a definition of a “core” or “main” idea of a text presented one of the challenges of our work with the writers. Each summary is supposed to be a cohesive standalone text that could be read and understood on its own.

The fact that the source documents represent different domains poses another challenge for the writers: they need to possess some knowledge and expertise in each of these. The documents are split into sections and paragraphs, and the writers are asked to keep the flow of the section/paragraph structure of the source text, trying to summarize it from top to bottom. However, we emphasize that the sentence by sentence summarization is exactly what we do not need, and the summaries need to be abstractive rather than extractive, which means copying the source text is strongly discouraged.

In addition, the writers are asked to provide a set of questions and answers for each long document. They need to compose at least 13--15 questions per 5,000 words. The answers are supposed to cover the points reflected in the summaries. We instruct the writers to produce open-ended, complex questions, which provide a good baseline for testing reasoning. For tracking and alignment purposes, each paragraph in each source document is given a number.  The writers are asked to specify in which paragraph each answer can be found. They are also asked to indicate which of the summaries provides the answer to each question.

Besides the general guidance, we discovered that working with conversational content needed additional clarifications, so we prepared an additional document for working specifically with long-form text that contains conversations (such as plays or screenplays).

\subsection{Automatic output and postprocessing}
\label{sec:generation}

We want to understand how current state-of-the-art models perform on our new benchmark, both on the capability of comprehension a very long context and on the generation of long outputs. We conduct the automatic abstractive summarization for the former and summary expansion for the latter. We give details on the tasks below.

\paragraph{Gradual Summarization.} 
We prompt the models with the human guidelines with a slight adaptation to be LLM friendly. We input the entire document without paragraph splitting. To give a fair evaluation, we prompt all LLMs in the zero-shot setting. To control the length of the LLM output, we have added additional instructions with the upper and lower bounds of the permissible words. For example, to ask the model to generate a summary of the R\% length of the source text, the prompt contains \texttt{"Make sure the summary has \{y\}  words or less.....Please write at least \{x\} words""""
}
, where \texttt{x} and \texttt{y} are determined per document with respect to the length and ratio \texttt{R}. In practice, we see that enforcing the length of the document right before and after the content block in the prompt gives consistent results. We give details of our summary prompts in the Appendix \ref{app:prompting}.

\paragraph{Summary expansion.} We customize prompts for each domain, plus the model-specific prompt templates. Similarly to the summarization task, the prompt contains instructions on the desired range of the generated text length. In addition, each prompt has instructions to guide the model in generating content of a certain quality (consistency, coherence, and keeping the main ideas in the summary). We prompt the model with specific formats for different domains as reported in the appendix~\ref{app:prompting}.

\paragraph{Automatic Paragraph Alignment (APA).}
We add this step to help human annotators evaluate the outputs that we are creating (as detailed in the next section \ref{subsec:humaneval}). The task of comparing long inputs and outputs creates a high cognitive load on human evaluators. To reduce it, we provide an approximate alignment between the input paragraphs and the segments of the output, taking advantage of the assumption that a summary usually follows the structure of the source document. First, we use dynamic programming to find a monotonic alignment path between input and output sentences that would maximize the sum of cosine similarities of the SONAR embeddings \citep{Duquenne:2023:sonar_arxiv} of the two sentences. An output sentence could be aligned with multiple consecutive input sentences, potentially from different paragraphs, but we assign it to a single input paragraph with which it is aligned the most frequently. Thus, each input paragraph gets aligned to a contiguous output segment (potentially empty) in a monotonic way. This alignment helps the annotators navigate the input and output documents jointly.

\subsection{Human Evaluation}
\label{subsec:humaneval}

To perform human evaluation on previously generated output, we design human evaluation guidelines inspired by previous works \citep{clark-etal-2023-seahorse,krishna-etal-2023-longeval,que2024hellobenchevaluatinglongtext} and fully reported in Appendices \ref{app:humaneval:summ} and \ref{app:isum_guidelines}. 

\paragraph{Human evaluation on gradual summaries.}

Before starting the evaluation, annotators are allowed to reject a task if the output text is gibberish or obviously of low quality.

The generated summaries are evaluated in two tasks. In Task 1, the annotators first read the source document and the three summaries and then rate the generated text in four aspects, including attribution, coverage of the main ideas, conciseness and readability (similar to the 'checklist' in HelloBench by \citet{que2024hellobenchevaluatinglongtext}). The annotators rate the summary on a 0-4 Likert scale and finally give an overall rating on a 0-10 Likert scale. Each summary receives its own separate set of scores. 

In Task 2, the annotators validate the QA sets that were previously created by human writers. For each question in the QA sets (13--15 questions and answers), the annotators are required to determine whether the content of the summary contains enough information to answer the question (i.e., the answer is directly stated, heavily implied or logically entailed in the summary). The annotators give a YES or NO to each QA pair. For each summary, the annotators validate the whole set of QA once.

The whole evaluation is referenceless, which means that the human written summaries are not shown to annotators, and that they only see a single set of summaries from one anonymous model output each time.

Both tasks 1 and 2 involve human judgment, and to reduce the bias, 3 sets of rating from random annotators are required for each generated output. The same guidance should be used for all different domains. Detailed evaluation guidelines are included in the Appendix \ref{app:humaneval:summ}.

\paragraph{Human evaluation on summary expansion.}

We use the same format as the previous summarization evaluation tasks and integrate some of the questions from Story Plot Generator \citep{zhu2023endtoendstoryplotgenerator} and HelloBench \citep{que2024hellobenchevaluatinglongtext}. For task 1, the annotators read the source summary and the generated long-form output, rate the output on 6 aspects, including the coverage of main core ideas, cohesion, richness in details, creativity, non-repetitiveness, and interest, and give an overall rating at the end. In task 2, they validate the QA set with the generated long-form text. Each output is evaluated separately without reference, and three sets of random annotation ratings are required. Detailed evaluation guidelines are included in the Appendix \ref{app:isum_guidelines}.

\paragraph{Evaluation statistics.} 

Summaries and summary expansions are each evaluated separately. For the evaluation of generated summaries, 252 documents from all domains are used as the source to generate the summaries (with 2 documents being excluded during the process). The summaries are generated using three different models (as reported in Section \ref{sec:experiments} and chosen to represent close and open models of different sizes): GPT-4o-mini-64k \citep{openai2024gpt4technicalreport}, \llama 3.1-70B, and \llama 3.1-8B \citep{dubey2024llama3herdmodels}. This results in 756 outputs and, along with 252 sets of human-written summaries, creates a dataset of 1,000 document-summary pairs for evaluation. A vendor sources 287 annotators, who are required (1) to be native speakers of English and (2) to hold a language-related degree. These annotators are selected from a pool that is different from that of the summary writers, ensuring that they have no prior knowledge of the source documents or the written summaries. Tasks are randomly assigned to annotators until every set of generated output receives three complete annotations. A limit of 10 evaluations for each model is set per annotator to mitigate biases in the results.



For evaluation of generated summary expansions, only a subset of data is selected,\footnote{We exclude domains with high density of factual information, because we believe they are less appropriate for the summary expansion task. The problem here is potential hallucination of factual information (e.g. \textit{Berlin is the capital of France}), which can be detrimental for real-life use cases but is out of scope of our evaluation that does not check for extrinsic factuality but only for faithfulness to the source text.} including SummScreen, BookSum, SQuality and FacetSum (102 source documents in total). The expansions are generated with the same models as previously (GPT-4o-mini, \llama 3.1-70B, and \llama 3.1-8B), resulting in 306 long-form outputs. Ten experienced data analysts are selected to conduct the evaluation. Similarly to the evaluation of summaries, the tasks are assigned randomly until every long form output receives 3 complete annotations.


\subsection{Automatic evaluation}

The summarization outputs are typically evaluated by computing ROUGE scores \citep{lin2004rouge} with respect to a reference. However, this approach is not sufficient for at least three reasons \citep{schluter2017limits}: it depends too much on the reference, it offers only a comparison at the surface level, and it does not explain why a summary is good or bad. Thus, we compute several other reference-free metrics, each targeting a specific aspect of summarization quality (over the 6 aspects introduced in the SEAHORSE dataset by \citet{clark-etal-2023-seahorse}). For each aspect, we tried to pick a metric that is focused on this specific aspect, preferring ones that are computationally transparent and that have already been implemented elsewhere: 

\begin{enumerate}[noitemsep,nolistsep]
    \item \textbf{Repetitiveness}: how much the summary repeats the same phrases. We report the count of all the word n-grams ($n\in\{1,2,3\}$) in the summary, divided by the count of such unique n-grams (REP-3) \citep{welleck2019neural}.
    \item \textbf{Fluency}: how grammatical the text is. We report the average probability of a summary sentence being grammatical (or linguistically acceptable in the Chomskyan sense of the term) computed with a CoLA classifier \citep{sent_fluency_style20}. 
    \item{\textbf{Coherence}}: how similar are the sentences in the generated texts to each other. COH-2 averages similarity of the neighboring-over-one sentences in the embedding space~\citep{parola2023speech}.
    \item \textbf{Attribution}: how much of the summary is directly attributable to the source (something like ``precision'' of the ideas in the summary). 
    The average score of SEAHORSE Q4 (SH-4) model evaluates attribution \citep{clark-etal-2023-seahorse}.   
    \item \textbf{Coverage of the source}: how much of the source is reflected in the summary. 
    The average SEAHORSE Q5 (SH-5) score reports this aspect \citep{clark-etal-2023-seahorse}. 
\item{\textbf{Overall}} In order to evaluate the overall quality of the text, we use two metrics: 
    \\ (1) For summarization, we aggregate a score (AVG) from the above metrics, namely, averaging $-REP\text{-}3, CoLA, COH\text{-}2, SH\text{-}4, SH\text{-}5$. For summary expansion, the aggregated score (AVG) is the average score of $REP\text{-}3, CoLA, COH\text{-}2$. Note that $REP\text{-}3$ is negated to make the score monotonic. Also, for the summary expansion, $REP\text{-}3$ increases the value over the length of output, so the factor $0.2$ is empirically set to normalize the value on the summary expansion task 20\%. 
    \\ (2) We use HelloEval (HE) score \citep{que2024hellobenchevaluatinglongtext}: an LLM-as-judge model with various checklists trained w.r.t. human evaluation.

\end{enumerate}

Selection of these metrics from a larger candidate set was partially motivated by their correlations with human annotations from \citet{clark-etal-2023-seahorse}, described in Appendix \ref{app:early_metrics_corrs}. Table \ref{tab:metrics} in Appendix \ref{app:eval} summarizes the list of metrics.

For the SEAHORSE scores, we had to feed the whole source text to a transformer model, which was neither feasible computationally with long context inputs nor made sense given the relatively short-form training data of those models. To bypass this problem, we segment sources and summaries into aligned fragments (using a modification of the alignment algorithm in Section \ref{sec:generation}) with at most 50 sentences on the source side and compute model-based metrics for the fragment pairs.

\subsection{Data Statistics}

LCFO covers 7 domains sourced from 10 datasets with an average document length of 5k words. 
Table \ref{tab:dataset-fields} contains the distribution of the LCFO dataset in subsets and domains, as well as the average word length of the documents. More details are reported in the Appendix \ref{app:data}.

\begin{table}[ht!]
\centering
\scriptsize
\begin{tabular}{llp{0.5\linewidth}l}
\hline
\textsc{dataset} & \textsc{N} & \textsc{domain} & \textsc{len}  \\ \hline
\addlinespace[0.3em]
LexGLUE & 25 & Legal: supreme court opinions & 4953 \\ \addlinespace[0.3em]
BookSum & 27 & Literature: books, novel, action & 4114  \\ \addlinespace[0.3em]
SQuality & 25 & Literature: stories & 4856  \\ \addlinespace[0.3em]
FacetSum & 25 & Scientific: journal articles on various domains & 4904  \\ \addlinespace[0.3em]
JRC-Acquis & 25 & Legal: legislative text of the European Union & 4825  \\ \addlinespace[0.3em]
MultiUN & 25 & Political: UN docs & 4539  \\ \addlinespace[0.3em]
Wikipedia & 25 & Wikipedia: 22 docs on biomedicine & 5266  \\ \addlinespace[0.3em]
GovReport & 25 & Political: Congressional Research Service and US Government Accountability Office & 5078  \\ \addlinespace[0.3em]
Summscreen & 25 & Conversational: TV series transcript & 5030  \\ \addlinespace[0.3em]
Seahorse & 25 & News: English BBC news & 4576  \\ \addlinespace[0.3em]\hline \addlinespace[0.3em]
Total & 252 & Average word count & 4814  \\ \addlinespace[0.3em]\hline

\end{tabular}
\caption{LCFO Summary: domains and statistics (number of documents \textsc{N} and average  length in words \textsc{len})}
\label{tab:dataset-fields}
\end{table}


\section{Experiments}
\label{sec:experiments}

\paragraph{Settings.}
We experimented with closed and open LLMs. We chose GPT-4o-mini-64k\footnote{\url{https://openai.com/index/gpt-4o-mini-advancing-cost-efficient-intelligence/}} for the closed model and \llama 3.1-70B~\citep{dubey2024llama} for the open-source one. 
For summarization, we ran the model with all length ratios (5\%, 10\%, 20\%), while for summary expansion, we only expanded the summaries 20\% to the full document. 
We also performed a postprocessing step to filter the templated response such as \texttt{``**Summary**''}, \texttt{``Here is the summary:''}, etc.

\paragraph{Summarization results.}
Table~\ref{tabl:summarization} shows the general results of the selected models at different levels of gradual summarization. Results broken down by domains are reported in the Appendix \ref{app:detailedresults}. Note that LLMs tend to perform similarly regardless of the length of the output in terms of human scores. This is not the case for humans that show to lag behind when performing short summaries. The best results are consistently achieved with GPT-4o-mini and are consistent with previous research findings \citep{que2024hellobenchevaluatinglongtext}. This model even surpasses human-level quality in short summaries. This may be explained by humans tending to perform worse when summarizing short documents and better when summarizing long ones.

\begin{table*}[ht!]
\centering
\scriptsize
\sisetup{table-format = 3.2}
\resizebox{\textwidth}{!}{%
\begin{tabular}{@{}p{2.2cm}cccccccccc@{}}
\toprule
Output & \multicolumn{1}{l}{\textsc{R-L(\(\uparrow\))}} & \textsc{REP-3(\(\downarrow\))} & \textsc{CoLA\(\uparrow\)} & \textsc{COH-2\(\uparrow\)} & \textsc{SH-4\(\uparrow\)} & \textsc{SH-5\(\uparrow\)} & \textsc{AVG\(\uparrow\)}  & \textsc{HE\(\uparrow\)} 
& Hum\(\uparrow\) \\
\midrule
\addlinespace[0.3em]
\rowcol\multicolumn{11}{c}{\textsc{LCFO.5\%}} \\
\addlinespace[0.5em]

Human & n/a &\textbf{0.308}  &0.941 & 0.809 & 0.644 & 0.387& 0.494  & 52.195 & 6.61 \\
\addlinespace[0.3em]
GPT-4o-mini & 0.331 & {0.328} & 0.968 & 0.719 & \textbf{0.635} & \textbf{0.487} & \textbf{0.496} & \textbf{76.917}   & \textbf{7.25}\\
\addlinespace[0.3em]

\llama 3.1-70B & 0.384 & 0.383 & 0.965 & 0.861 & 0.622 & 0.377 & 0.488&72.468   & 6.27\\
\addlinespace[0.3em]

\llama 3.1-8B & 0.377 & 0.411 & \textbf{0.969} & \textbf{0.865} & 0.618 & 0.372 & 0.482 &  63.894  & 6.32  \\
\addlinespace[0.5em]


\midrule
\addlinespace[0.3em]
\rowcol\multicolumn{11}{c}{\textsc{LCFO.10\%}} \\
\addlinespace[0.5em]

Human & n/a & \textbf{0.395 } & 0.945 &  0.816  & \textbf{0.661}&   0.416 &  \textbf{0.489} & 64.688   & 7.44 \\

\addlinespace[0.3em]
GPT-4o-mini & 0.385 & {0.404} & \textbf{0.964} & 0.695 & 0.621 & \textbf{0.471} & 0.469 & \textbf{77.863}  & \textbf{7.50}\\
\addlinespace[0.3em]

\llama 3.1-70B & 0.434 & 0.515 & 0.944 & \textbf{0.860} & 0.614 & 0.369 & 0.454 &  72.497   & 6.42\\
\addlinespace[0.3em]

\llama 3.1-8B & 0.429 & 0.534 & 0.963 & 0.858 & 0.612 & 0.366 & 0.453 &  59.385  & 6.63  \\
\addlinespace[0.3em]


\midrule
\addlinespace[0.3em]
\rowcol\multicolumn{11}{c}{\textsc{LCFO.20\%}} \\
\addlinespace[0.5em]

Human & n/a &\textbf{0.244} &  0.938 &  0.805 &  0.615  & 0.357 &  \textbf{0.494}& 69.745
& \textbf{7.78} \\
\addlinespace[0.3em]

GPT-4o-mini & 0.445 & {0.497} & \textbf{0.961} & 0.673 & \textbf{0.616} & \textbf{0.464} & 0.443 & \textbf{76.706}  & 7.52 \\
\addlinespace[0.3em]

\llama 3.1-70B & 0.467 & 0.631 & 0.928 & 0.860 & 0.596 & 0.357 & 0.422 &  71.603   & 6.32\\
\addlinespace[0.3em]

\llama 3.1-8B & 0.469 & 0.647 & 0.956 & \textbf{0.861} & 0.594 & 0.370 & 0.427 &  51.015 & 6.60  \\
\addlinespace[0.3em]

\addlinespace[0.5em]

\bottomrule
\end{tabular}%
}
\caption{Performance on the summarization task}
\label{tabl:summarization}
\end{table*}

\paragraph{Summary expansion results.}
Table~\ref{tabl:isummarization} shows the overall results on the summary expansion task by a factor of 5, giving the 20\% summary input. The performance of models is not coherent across metrics that look only at the output (i.e.\%WC, REP-3, CoLA, COH-2, and AVG). In terms of HE and coherently with the human evaluation results, GPT-4o-mini is the best performing model. 
Additionally, we report the results of other combinations, for example, expanding by larger factors (10 and 20) giving 10\% summary input and giving 5\% summary input, respectively, in Appendix \ref{app:detailedresults}.

\begin{table*}[ht!]
\centering
\scriptsize
\sisetup{table-format = 3.2}
\begin{tabular}{@{}p{2.2cm}cccccccc@{}}
\toprule
Output & \%WC & \textsc{REP-3(\(\downarrow\))} & \textsc{CoLA\(\uparrow\)} & \textsc{COH-2\(\uparrow\)} & \textsc{AVG\(\uparrow\)}  & \textsc{HE\(\uparrow\)} 
& Hum\(\uparrow\) \\
\midrule

GPT-4o-mini & \textbf{1.931} & 0.707 & \textbf{0.913} & 0.609 & 0.460 & \textbf{70.896} & \textbf{6.431} \\
\addlinespace[0.3em]

\llama 3.1-70B & 1.058 & \textbf{0.680} & 0.877 & 0.750 & 0.497 & 39.199  & 4.469 \\
\addlinespace[0.3em]

\llama 3.1-8B & 1.187 & 0.809 & 0.903 & \textbf{0.779} & \textbf{0.507} & 38.416 & 4.801\\

\addlinespace[0.5em]

\bottomrule
\end{tabular}%
\caption{Performance on the summary expansion task by a factor of 5, giving the 20\% summary input.}
\label{tabl:isummarization}
\end{table*}


When comparing across tasks (i.e. summarizing to 20\% or doing summary expansion from 20\% by a factor of 5), the results show better performance in the former.  It is expected that summary expansion is a more challenging task across domains and all models. Current models struggle with this task. If we compare HE, the deltas in the same model vary from 6\% for GPT-4o-mini to $\approx$ 30\% for \llama 3.1-70B. When comparing output-based metrics, there are discrepancies in conclusions (i.e., \llama 3.1-8B better than 70B model). However, HE is still worse for the 8B model. This may indicate that selected output-based quality metrics are less reliable than the HE score (see the analysis below for metrics evaluation).  

\begin{table*}[ht!]
\centering
\fontsize{9}{9}\selectfont
\sisetup{table-format = 3}
\begin{tabular}{@{}ccccccc@{}}
\toprule
&\textsc{R-L}& \textsc{CoLA} & \textsc{SH-4} & \textsc{SH-5} & \textsc{AVG}  & \textsc{HE}  \\ 
\midrule \addlinespace[0.3em]
S & 0.196 (0.065) & 0.595 (6.337e-10)& 0.616 (1.005e-10)  & 0.445 (1.105e-5)  &  0.159 (0.135) & 0.428 (2.591e-05) \\ \addlinespace[0.3em]\midrule
SE & n/a & n/a  & n/a & n/a  & 0.285 (3.646e-05) & 0.405 (1.957e-09) \\
\bottomrule
\end{tabular}%
\caption{Spearman correlation coefficients (and p-value) for various aspects and overall scores between automatic metrics and human evaluation for summarization (S) and summary expansion (SE). For the former, we show correlations between CoLA and Human evaluation Q2d; SH-4 and Human evaluation Q2a; SH-5 and Human evaluation Q2b and R-L/AVG/HE and Human evaluation Q3 (appendix \ref{app:humaneval:summ}). For the latter, we show correlations between AVG/HE and Human evaluation Q3 (appendix \ref{app:isum_guidelines}).}
\label{tab:dataset-fields-isum}
\end{table*}

\paragraph{Metrics evaluation.} In our study, we consider human evaluation, conducted according to the guidelines outlined in Appendices \ref{app:humaneval:summ} and \ref{app:isum_guidelines}, as the definitive measure of the overall quality score, as well as the scores for individual quality aspects such as coverage and attribution. To mitigate potential biases among the annotators, we calculate the average of three annotations for each task. 
Table \ref{tab:dataset-fields-isum} presents the Spearman correlation coefficients for various aspects and overall scores, comparing automatic metrics and human evaluations for summarization and summary expansion, respectively. A higher Spearman correlation coefficient signifies a stronger correlation between the automatic metrics and human annotation. The metric that shows the highest correlation with human annotation corresponds to SH-4, which measures attribution. When comparing metrics that measure overall performance,  we observe that R-L is not very good at correlation, but it may also be due to the fact that our task is not the best suited to use human references. HE is the one with the highest correlation. AVG low-correlation (both in S and SE) may be explained by the fact that individual averaged metrics are not very good or they cover more specific aspects which may not end capturing the overall performance. This low correlation for R-L and AVG can explain the discrepancy observed in the model ranking (specially between \llama 3.1-70B and 8B in Tables \ref{tabl:summarization} and \ref{tabl:isummarization}. Correlations are low in all cases, which shows the difficulty of the evaluation. Beyond the challenge of automatizing it, we should add the fact that humans struggle in generating short summaries, which may imply that humans also struggle in evaluating them.

\section{Related Work}

Related work on long context and long form output comes in many flavors. We cover a summary on long context and long-form output datasets.

\paragraph{Long-context datasets.} Infinite length datasets such as NIAH, RULER \citep{hsieh2024rulerwhatsrealcontext} work with distracting information. Finite-length nondistractive-based datasets include: Multi-LexSum \citep{10.5555/3600270.3601226} is a collection of 9,280 expert-authored summaries drawn from single domain (Civil Rights Litigation Clearinghouse) writing the length of the source documents often exceeds two hundred pages per case and summaries are presented in two-sizes: a short and longer version; Longbench \citep{bai-etal-2024-longbench}  and Marathon \citep{zhang-etal-2024-marathon} that includes tasks with 5--25k context and, more recently, \cite{kwan-etal-2024-m4le} build a dataset up to 8k tokens context length to evaluate LLMs’ long-context understanding across five key abilities: understanding of single or multiple relevant spans in long contexts based on explicit or semantic hints, and global context understanding. Loong is a multi-document QA dataset up to 200k context to assess RAG abilities. HelloBench \citep{que2024hellobenchevaluatinglongtext} includes summarization of a selection of long-input documents (3k to 6k word length).

\paragraph{Long-form output datasets.} There is a lack of reference-based datasets on long form output. However, there are datasets that study prompting of different long-form generation; e.g., StoryGen \citep{zhu2023endtoendstoryplotgenerator} includes prompts to generate stories, and HelloBench is one of the most diverse long form generation benchmarks including stories, screenplays, keyword writing.

Our contribution on datasets involve the manual collection of 3-length summaries from long input documents. This collection also includes abstractive QA (non-multiple choice) to test comprehension.
Our contribution on metrics involves new human evaluation protocols on summarization and summary expansion, as well as annotations to develop supervised metrics on long-form outputs.

\section{Conclusions}

 \lcfo{} provides gradual summaries references from 5k input documents with QA pairs for each of the documents and summaries. Additionally, we provide human evaluation of human and model-generated summaries and model-generated summary expansions. Overall, \lcfo{} 
 enables the evaluation of several tasks and metrics in the setting of long-context input documents and long-form output. While the main contribution of this paper is to present the freely available \lcfo{} dataset\footnote{Available at \url{https://huggingface.co/datasets/facebook/LCFO}},
 we also evaluate model and human outputs, showing that LLMs are capable of surpassing human results when producing short summaries. Current evaluation results question the usefulness of manually generating human references for short summarization of long documents.
To confirm this, as further work, we plan to exploit the capabilities of LCFO by using QA as part of automatic evaluation (i.e. scoring how many questions are correctly answered in model-generated summaries).

\section*{Limitations and Ethical considerations}

\paragraph{Data contamination.} Source documents may exist in the training data of the models, therefore, generation may be at risk. To mitigate this, we prioritize recent documents, since this is not enough, we annotate the correspondence of sections in summarization versions, so that we can generate only portions of the document. Therefore, if the model uses internal knowledge, we can quantify by spotting details from other sections.

\paragraph{Experiments.} The experimental options that \lcfo{} offers are much larger than the ones we explore in this paper. Also, the dataset can be easily expanded to have more summary references by matching with existing summaries in some of the domains. QA pairs have not been used in the paper but this is designed (but not limited) to serve for doing reading comprehension and/or for creating an evaluation metric.

\paragraph{Metrics.} Summarization is a generative task with very diverse aspects of quality, and no single automatic evaluation metric captures them all adequately. To compensate for this, we report multiple evaluation metrics, but still, some of them are not well established; for example, there is no single metric of longform text coherence that the summarization community agrees upon. By providing the results of the human evaluation, we hope to help the community develop and validate better automatic evaluation metrics in the future.

\paragraph{Computing} In terms of computing, evaluating LLMs on \lcfo{} benchmark require larger memory due to both its big context size and the long-form output (should the models be capable to it). In case of \llama, we used 1 NVIDIA GPU A100 80 GB for the 8B model, and 8 GPUs for the 70B model. The resource was shared with the loading of scoring models (SH, CoLA) as well. Each evaluation run over 10 domains takes 90 minutes, including the computation of all the scores except HelloEval (where the computing time depends on the external availability of GPT-4 endpoint deployment).

\paragraph{Annotations} Annotators were paid a fair rate.  Each of the annotators signed a consent form agreeing on the dataset and its usage that they were participating in.

\section*{Acknowledgements}

This paper is part of the LCM project\footnote{https://github.com/facebookresearch/large$\_$concept$\_$models} and authors would like to thank the entire LCM team for the fruitful discussions.

\bibliography{custom}

\appendix

\section{Automatic Paragraph Splitting Details}
\label{app:paragraph}

As part of the guidelines for summarization, annotators are instructed to read long documents from different domains and mentally distill key points into paragraphs and then into a cohesive document summary. This process requires the source text to be logically segmented into well-structured paragraphs that facilitate comprehension and synthesis.

During pilot studies, it became evident that the quality of the initial paragraph segmentation significantly impacted annotation outcomes. Poorly segmented paragraphs increased cognitive load and risked misinterpretation, while cohesive and logically structured paragraphs improved annotation consistency and efficiency.

Given the variability in formats and structures across datasets, a uniform approach to paragraph splitting was not feasible. Some datasets provided explicit structural markers (e.g., new lines, section headers), while others required more algorithmic intervention, such as employing the Segment Anything Text (SaT-l3) model. Furthermore, the SaT-l3 model's performance varied across text types, necessitating dataset-specific thresholds and post-processing techniques to optimize paragraph segmentation.

This section outlines the tailored methodologies applied to each dataset in the LCFO corpus, highlighting how their unique characteristics were addressed to produce high-quality, preprocessed documents for annotation.

\paragraph{LexGLUE}
Paragraphs were split using double newlines as separators, preserving the inherent paragraph structure in the dataset.

\paragraph{BookSum and SQualITY}
Lines were joined with a blank space to create continuous text blocks.
Sentences and paragraphs were split using the SaT-l3 model with a threshold of 0.8, producing lists of sentences grouped into paragraphs.
Paragraphs exceeding 3,000 characters were further split using the SaT-l3 model with a stricter threshold of 0.4.
Consecutive short paragraphs (fewer than 2 sentences or under 400 characters) were merged to ensure coherence, especially for dialogue-heavy sections.

\paragraph{JRC-Acquis} 

Lines were joined with a blank space to preserve the flow of text.
Sentences and paragraphs were split using the SaT-l3 model with a standard threshold of 0.5.
Consecutive paragraphs containing fewer than 2 sentences were merged.
Sections and subsections were extracted from paragraph beginnings using the dataset’s consistent numbered format (e.g., 1.1.2), serving as structural indicators.

\paragraph{MultiUN}

Lines were joined using blank spaces to form initial text blocks.
Sentences and paragraphs were split using the SaT-l3 model with a threshold of 0.5.
Short consecutive paragraphs (fewer than 2 sentences each, and up to 20 sentences total) were merged to improve readability and flow.

\paragraph{Wikipedia}

Original paragraphs were identified using empty lines (meaning double newline in the original text), which appeared as blank lines or in the CSV format.
Long paragraphs (over 500 tokens) were split further using the SaT-l3 model with a threshold of 0.5 to improve segmentation accuracy for longer text units.

\paragraph{GovReport}
Same as LexGLUE, paragraphs were split using double newlines as separators.

\paragraph{Summscreen}
Initial paragraph segmentation was based on scene indicators ([SCENE-BREAK]) in the transcripts. However, this often resulted in excessively long paragraphs, with some documents containing only one or two paragraphs.
Text formatting issues, such as double spaces in punctuation (e.g., " . "), were corrected to align with the SaT-l3 model’s sensitivity.
Long paragraphs exceeding 3,000 characters were re-segmented using the SaT-l3 model with a threshold of 0.9.
Short consecutive paragraphs containing only one sentence were merged to form cohesive segments.

\section{Data Details}
\label{app:data}

\paragraph{Our data collection}

\begin{itemize}
    \item  100\% human annotated (no LLM pre-selection)
\item 7 domains (political, wikipedia, scientific, literature, conversational, legal
\item 252 Source Documents (~5k)
\item 4 lengths of the same Source Document ($\approx$5k, $\approx$1k, $\approx$500, $\approx$250 words)
\item 13-15 QA on each Long Context Source Document
\item Annotation on the presence of these QA on each of the summaries 

\item Human evaluation of automatic and manual summaries
\item Human evaluation of summary expansion
\end{itemize}








\section{Summary evaluation}
\label{app:eval}

Table \ref{tab:metrics} summarises the metrics used to evaluate.

\begin{table*}[ht!]
\centering
\scriptsize
\begin{tabular}{lp{2.5cm}lp{4.5cm}p{3.5cm}lp{2.5cm}}
\toprule
{\bf Task} & {\bf Area} & {\bf Metric} & {\bf Description} & {\bf Reference}  \\  \midrule \addlinespace[0.3em]
{\bf Sum} & Target similarity & R-L & ROUGE-L (longest common subsequence) & ~\citet{lin2004rouge}  \\ \addlinespace[0.3em]
{\bf Sum/SumExp}& Grammaticality & REP-3 & Portion of duplicated N-grams (N=4) & \citet{welleck2019neural} \\\addlinespace[0.3em]
{\bf Sum/SumExp}& Fluency & CoLA & Sentence fluency classifier score & ~\citet{sent_fluency_style20} \\\addlinespace[0.3em]
{\bf Sum/SumExp}& Coherence & COH-2 & 2nd-order word-level coherence score & ~\citet{parola2023speech} \\\addlinespace[0.3em]
{\bf Sum}& Attribution & SH-4 & Seahorse-Large-Q4 score & \citet{clark-etal-2023-seahorse} \\\addlinespace[0.3em]
{\bf Sum}& Semantic coverage & SH-5 & Seahorse-Large-Q5 coverage score & \citet{clark-etal-2023-seahorse} \\\addlinespace[0.3em]
{\bf SumExp} & Word Count & WC &  &   \\ \addlinespace[0.3em]
{\bf Sum/SumExp}& Overall & AVG & Empirical average of metrics & \\\addlinespace[0.3em]
{\bf Sum/SumExp}& Overall & HE & HelloEval score &\citet{que2024hellobenchevaluatinglongtext} \\\addlinespace[0.3em]
\bottomrule
\end{tabular}
\caption{\label{tab:metrics}Summary of automatic metrics used in different tasks. }
\end{table*}

\section{Selection of automatic metrics}
\label{app:early_metrics_corrs}

At an early stage of our work (before collecting human annotations), we came up with a list of candidate metrics of summarization quality. We provide it in Table \ref{tab:metrics_early_candidates}.

To select a single metric per aspect, we analyzed their Spearman correlations with human annotations of the English subset of SEAHORSE \citep{clark-etal-2023-seahorse}, focusing on Questions 2 to 6: absence of repetitions, grammatical correctness, attribution to the source, coverage of the source, and conciseness (which also reflects overall quality). We did not work with Question 1 (comprehensibility), because our metrics focus on inter-sentential comprehensibility, whereas 66\% of SEAHORSE summaries consist of a single sentence.

We report the resulting correlations, grouped by the three English SEAHORSE subsets, in Table \ref{tab:metrics_early_corrs}. One can see that the metrics we selected for summarization correlate reasonably with their corresponding aspects across all 3 subsets most of the time\footnote{An exception is the grammaticality/fluency aspect (Question 2) on the Wikihow subset, where no metric demonstrated adequate correlation with human annotations.}, which justifies our choice of them.

\begin{table*}[ht]
\renewcommand{\tabcolsep}{2pt}  
\centering
\scriptsize
\begin{tabular}{llp{0.66\linewidth}}
\toprule
Metric & Short name & Implementation \\
\midrule
ref\_rouge\_fmeasure              & R-L   & ROUGE-L F1 score, computed with the \texttt{rouge-score} Python package (in the \texttt{rougeLsum} mode) \\
ref\_rouge\_recall                &       &  ROUGE-L recall (see above) \\
ref\_rouge\_precision             &       &  ROUGE-L precision (see above) \\
char\_len\_ratio                  &       &  Summary-to-source lengths ratio, in characters  \\
sent\_len\_ratio                  &       & Summary-to-source lengths ratio, in sentences  \\
word\_len\_ratio                  &       & Summary-to-source lengths ratio, in words \\
td\_coherence\_1st                &       & 1st order coherence, computed with the \texttt{textdescriptives} Python package \\
td\_coherence\_2nd                & COH-2 & 2nd order coherence, computed with the \texttt{textdescriptives} Python package \\
td\_flesch\_kincaid\_grade        &       & Flesch-Kincaid grade readability score (\texttt{textdescriptives}) \\
td\_gunning\_fog                  &       & Gunning-Fog readability score (\texttt{textdescriptives}) \\
td\_dependency\_distance\_mean    &       & Mean distance of syntactic dependencies (\texttt{textdescriptives}) \\
word\_ngram\_src\_overlap         &       & Fraction of source word n-grams (1 to 3) that appear in the summary \\
word\_ngram\_repetition\_rate     & REP-3 & Fraction of summary word n-grams (1 to 3) that appear more than once \\
src\_rouge\_recall                &       & ROUGE-2 recall of the summary w.r.t. the source \\
src\_rouge\_f1                    &       &  ROUGE-2 F1 score of the summary w.r.t. the source \\
src\_rouge\_precision             &       & ROUGE-2 precision of the summary w.r.t. the source \\
tgt\_fluency                      & CoLA  &  Mean predicted probability of the sentence being linguistically acceptable, as per the model from \citet{sent_fluency_style20} \\
src\_fluency\_diff                &       &  Difference of mean predicted probabilities of being acceptable in the summary and in the source \\
mean\_self\_sonar\_sim            &       & Mean cosine similarity SONAR embeddings of each summary sentence to its most similar summary sentence \\
mean\_src\_sonar\_sim             &       & Mean cosine similarity SONAR embeddings of each summary sentence to its most similar source sentence \\
mean\_src\_coverage\_sonar\_sim   &       & Mean cosine similarity SONAR embeddings of each source sentence to its most similar summary sentence \\
mean\_monotonic\_src\_sonar\_sim  &       & Mean cosine similarity SONAR embeddings of each summary sentence to its monotonically aligned source sentence  \\
sonar\_sim\_monotonicity          &       & Ratio of \texttt{mean\_monotonic\_src\_sonar\_sim} to \texttt{mean\_src\_sonar\_sim}                            \\
p\_entail\_full                   &       & Probability of the summary being entailed by the source, predicted with \texttt{tasksource/deberta-small-long-nli} model  \\
p\_noncontradict\_full            &       & Probability of the summary not contradicting the source, predicted with \texttt{tasksource/deberta-small-long-nli} model \\
smollm\_reconstruct\_loss         &       & Cross-entropy loss of the source given the summary, computed with the \texttt{HuggingFaceTB/SmolLM-360M-Instruct} model \\
seahorse\_q4                      & SH-4  & Probability of the ``Yes'' response extracted from the SEAHORSE Q4 model (\texttt{google/seahorse-large-q4}) \\
seahorse\_q5                      & SH-5  & Probability of the ``Yes'' response extracted from the SEAHORSE Q5 model  \\
seahorse\_q6                      &       & Probability of the ``Yes'' response extracted from the SEAHORSE Q6 model  \\
\bottomrule
\end{tabular}
\caption{\label{tab:metrics_early_candidates} The automatic metrics we considered as candidates, along with their implementation details.}
\end{table*}

\begin{table*}
\renewcommand{\tabcolsep}{2pt}  
\centering
\scriptsize
\begin{tabular}{l|ccccc|ccccc|ccccc}
\toprule
SEAHORSE subset & \multicolumn{5}{c}{wiki} & \multicolumn{5}{c}{xlsum} & \multicolumn{5}{c}{xsum} \\
Quality aspect & q2 & q3 & q4 & q5 & q6 & q2 & q3 & q4 & q5 & q6 & q2 & q3 & q4 & q5 & q6 \\
\midrule
ref\_rouge\_fmeasure (R-L) & {\cellcolor[HTML]{04588A}} \color[HTML]{F1F1F1} 0.33 & {\cellcolor[HTML]{76AAD0}} \color[HTML]{F1F1F1} 0.03 & {\cellcolor[HTML]{A9BFDC}} \color[HTML]{000000} 0.11 & {\cellcolor[HTML]{348EBF}} \color[HTML]{F1F1F1} 0.25 & {\cellcolor[HTML]{2786BB}} \color[HTML]{F1F1F1} 0.27 & {\cellcolor[HTML]{0C74B2}} \color[HTML]{F1F1F1} 0.17 & {\cellcolor[HTML]{4897C4}} \color[HTML]{F1F1F1} 0.12 & {\cellcolor[HTML]{549CC7}} \color[HTML]{F1F1F1} 0.12 & {\cellcolor[HTML]{04598C}} \color[HTML]{F1F1F1} 0.23 & {\cellcolor[HTML]{157AB5}} \color[HTML]{F1F1F1} 0.18 & {\cellcolor[HTML]{056DAB}} \color[HTML]{F1F1F1} 0.11 & {\cellcolor[HTML]{3991C1}} \color[HTML]{F1F1F1} 0.11 & {\cellcolor[HTML]{93B5D6}} \color[HTML]{000000} 0.13 & {\cellcolor[HTML]{529BC7}} \color[HTML]{F1F1F1} 0.16 & {\cellcolor[HTML]{76AAD0}} \color[HTML]{F1F1F1} 0.15 \\
ref\_rouge\_recall & {\cellcolor[HTML]{045C90}} \color[HTML]{F1F1F1} 0.31 & {\cellcolor[HTML]{8FB4D6}} \color[HTML]{000000} 0.02 & {\cellcolor[HTML]{8CB3D5}} \color[HTML]{000000} 0.15 & {\cellcolor[HTML]{2C89BD}} \color[HTML]{F1F1F1} 0.26 & {\cellcolor[HTML]{308CBE}} \color[HTML]{F1F1F1} 0.25 & {\cellcolor[HTML]{1278B4}} \color[HTML]{F1F1F1} 0.16 & {\cellcolor[HTML]{4C99C5}} \color[HTML]{F1F1F1} 0.11 & {\cellcolor[HTML]{67A4CC}} \color[HTML]{F1F1F1} 0.10 & {\cellcolor[HTML]{045483}} \color[HTML]{F1F1F1} 0.23 & {\cellcolor[HTML]{1E80B8}} \color[HTML]{F1F1F1} 0.18 & {\cellcolor[HTML]{056EAD}} \color[HTML]{F1F1F1} 0.11 & {\cellcolor[HTML]{60A1CA}} \color[HTML]{F1F1F1} 0.09 & {\cellcolor[HTML]{96B6D7}} \color[HTML]{000000} 0.12 & {\cellcolor[HTML]{4496C3}} \color[HTML]{F1F1F1} 0.17 & {\cellcolor[HTML]{79ABD0}} \color[HTML]{F1F1F1} 0.14 \\
ref\_rouge\_precision & {\cellcolor[HTML]{045F95}} \color[HTML]{F1F1F1} 0.29 & {\cellcolor[HTML]{99B8D8}} \color[HTML]{000000} 0.02 & {\cellcolor[HTML]{D0D1E6}} \color[HTML]{000000} 0.03 & {\cellcolor[HTML]{6FA7CE}} \color[HTML]{F1F1F1} 0.16 & {\cellcolor[HTML]{529BC7}} \color[HTML]{F1F1F1} 0.20 & {\cellcolor[HTML]{0C74B2}} \color[HTML]{F1F1F1} 0.17 & {\cellcolor[HTML]{4A98C5}} \color[HTML]{F1F1F1} 0.11 & {\cellcolor[HTML]{4295C3}} \color[HTML]{F1F1F1} 0.14 & {\cellcolor[HTML]{04649D}} \color[HTML]{F1F1F1} 0.20 & {\cellcolor[HTML]{157AB5}} \color[HTML]{F1F1F1} 0.18 & {\cellcolor[HTML]{056BA7}} \color[HTML]{F1F1F1} 0.12 & {\cellcolor[HTML]{4C99C5}} \color[HTML]{F1F1F1} 0.10 & {\cellcolor[HTML]{8CB3D5}} \color[HTML]{000000} 0.14 & {\cellcolor[HTML]{62A2CB}} \color[HTML]{F1F1F1} 0.14 & {\cellcolor[HTML]{78ABD0}} \color[HTML]{F1F1F1} 0.15 \\
char\_len\_ratio & {\cellcolor[HTML]{4496C3}} \color[HTML]{F1F1F1} 0.03 & {\cellcolor[HTML]{197DB7}} \color[HTML]{F1F1F1} 0.08 & {\cellcolor[HTML]{E9E5F1}} \color[HTML]{000000} -0.04 & {\cellcolor[HTML]{D8D7E9}} \color[HTML]{000000} -0.04 & {\cellcolor[HTML]{D7D6E9}} \color[HTML]{000000} -0.06 & {\cellcolor[HTML]{80AED2}} \color[HTML]{F1F1F1} 0.00 & {\cellcolor[HTML]{B8C6E0}} \color[HTML]{000000} 0.03 & {\cellcolor[HTML]{79ABD0}} \color[HTML]{F1F1F1} 0.09 & {\cellcolor[HTML]{0566A0}} \color[HTML]{F1F1F1} 0.20 & {\cellcolor[HTML]{4C99C5}} \color[HTML]{F1F1F1} 0.13 & {\cellcolor[HTML]{2F8BBE}} \color[HTML]{F1F1F1} 0.05 & {\cellcolor[HTML]{EBE6F2}} \color[HTML]{000000} -0.01 & {\cellcolor[HTML]{E6E2EF}} \color[HTML]{000000} -0.06 & {\cellcolor[HTML]{79ABD0}} \color[HTML]{F1F1F1} 0.12 & {\cellcolor[HTML]{C4CBE3}} \color[HTML]{000000} 0.03 \\
sent\_len\_ratio & {\cellcolor[HTML]{60A1CA}} \color[HTML]{F1F1F1} -0.02 & {\cellcolor[HTML]{023858}} \color[HTML]{F1F1F1} 0.14 & {\cellcolor[HTML]{EEE8F3}} \color[HTML]{000000} -0.06 & {\cellcolor[HTML]{D9D8EA}} \color[HTML]{000000} -0.05 & {\cellcolor[HTML]{D9D8EA}} \color[HTML]{000000} -0.07 & {\cellcolor[HTML]{9CB9D9}} \color[HTML]{000000} -0.04 & {\cellcolor[HTML]{E1DFED}} \color[HTML]{000000} -0.02 & {\cellcolor[HTML]{7BACD1}} \color[HTML]{F1F1F1} 0.08 & {\cellcolor[HTML]{358FC0}} \color[HTML]{F1F1F1} 0.13 & {\cellcolor[HTML]{81AED2}} \color[HTML]{F1F1F1} 0.08 & {\cellcolor[HTML]{1B7EB7}} \color[HTML]{F1F1F1} 0.08 & {\cellcolor[HTML]{EDE8F3}} \color[HTML]{000000} -0.02 & {\cellcolor[HTML]{E3E0EE}} \color[HTML]{000000} -0.05 & {\cellcolor[HTML]{B3C3DE}} \color[HTML]{000000} 0.05 & {\cellcolor[HTML]{D1D2E6}} \color[HTML]{000000} 0.01 \\
word\_len\_ratio & {\cellcolor[HTML]{4697C4}} \color[HTML]{F1F1F1} 0.03 & {\cellcolor[HTML]{1278B4}} \color[HTML]{F1F1F1} 0.08 & {\cellcolor[HTML]{ECE7F2}} \color[HTML]{000000} -0.05 & {\cellcolor[HTML]{DBDAEB}} \color[HTML]{000000} -0.05 & {\cellcolor[HTML]{D9D8EA}} \color[HTML]{000000} -0.07 & {\cellcolor[HTML]{8BB2D4}} \color[HTML]{000000} -0.02 & {\cellcolor[HTML]{B4C4DF}} \color[HTML]{000000} 0.03 & {\cellcolor[HTML]{86B0D3}} \color[HTML]{000000} 0.07 & {\cellcolor[HTML]{056FAE}} \color[HTML]{F1F1F1} 0.18 & {\cellcolor[HTML]{69A5CC}} \color[HTML]{F1F1F1} 0.11 & {\cellcolor[HTML]{2C89BD}} \color[HTML]{F1F1F1} 0.06 & {\cellcolor[HTML]{EDE7F2}} \color[HTML]{000000} -0.02 & {\cellcolor[HTML]{E9E5F1}} \color[HTML]{000000} -0.07 & {\cellcolor[HTML]{83AFD3}} \color[HTML]{F1F1F1} 0.11 & {\cellcolor[HTML]{CDD0E5}} \color[HTML]{000000} 0.02 \\
td\_flesch\_kincaid\_grade & {\cellcolor[HTML]{2F8BBE}} \color[HTML]{F1F1F1} 0.08 & {\cellcolor[HTML]{C5CCE3}} \color[HTML]{000000} -0.01 & {\cellcolor[HTML]{ACC0DD}} \color[HTML]{000000} 0.10 & {\cellcolor[HTML]{91B5D6}} \color[HTML]{000000} 0.11 & {\cellcolor[HTML]{88B1D4}} \color[HTML]{000000} 0.11 & {\cellcolor[HTML]{67A4CC}} \color[HTML]{F1F1F1} 0.04 & {\cellcolor[HTML]{88B1D4}} \color[HTML]{000000} 0.07 & {\cellcolor[HTML]{B8C6E0}} \color[HTML]{000000} 0.01 & {\cellcolor[HTML]{509AC6}} \color[HTML]{F1F1F1} 0.11 & {\cellcolor[HTML]{7EADD1}} \color[HTML]{F1F1F1} 0.09 & {\cellcolor[HTML]{589EC8}} \color[HTML]{F1F1F1} 0.01 & {\cellcolor[HTML]{DFDDEC}} \color[HTML]{000000} 0.00 & {\cellcolor[HTML]{A2BCDA}} \color[HTML]{000000} 0.10 & {\cellcolor[HTML]{73A9CF}} \color[HTML]{F1F1F1} 0.12 & {\cellcolor[HTML]{8CB3D5}} \color[HTML]{000000} 0.12 \\
td\_gunning\_fog & {\cellcolor[HTML]{1B7EB7}} \color[HTML]{F1F1F1} 0.14 & {\cellcolor[HTML]{E3E0EE}} \color[HTML]{000000} -0.03 & {\cellcolor[HTML]{9FBAD9}} \color[HTML]{000000} 0.12 & {\cellcolor[HTML]{7BACD1}} \color[HTML]{F1F1F1} 0.15 & {\cellcolor[HTML]{75A9CF}} \color[HTML]{F1F1F1} 0.15 & {\cellcolor[HTML]{71A8CE}} \color[HTML]{F1F1F1} 0.02 & {\cellcolor[HTML]{7EADD1}} \color[HTML]{F1F1F1} 0.08 & {\cellcolor[HTML]{C2CBE2}} \color[HTML]{000000} -0.00 & {\cellcolor[HTML]{7EADD1}} \color[HTML]{F1F1F1} 0.07 & {\cellcolor[HTML]{A2BCDA}} \color[HTML]{000000} 0.05 & {\cellcolor[HTML]{308CBE}} \color[HTML]{F1F1F1} 0.05 & {\cellcolor[HTML]{F2ECF5}} \color[HTML]{000000} -0.03 & {\cellcolor[HTML]{A4BCDA}} \color[HTML]{000000} 0.10 & {\cellcolor[HTML]{71A8CE}} \color[HTML]{F1F1F1} 0.13 & {\cellcolor[HTML]{84B0D3}} \color[HTML]{F1F1F1} 0.13 \\
td\_dependency\_distance\_mean & {\cellcolor[HTML]{348EBF}} \color[HTML]{F1F1F1} 0.07 & {\cellcolor[HTML]{F2ECF5}} \color[HTML]{000000} -0.05 & {\cellcolor[HTML]{DFDDEC}} \color[HTML]{000000} -0.01 & {\cellcolor[HTML]{CDD0E5}} \color[HTML]{000000} -0.01 & {\cellcolor[HTML]{CED0E6}} \color[HTML]{000000} -0.04 & {\cellcolor[HTML]{3F93C2}} \color[HTML]{F1F1F1} 0.09 & {\cellcolor[HTML]{B8C6E0}} \color[HTML]{000000} 0.03 & {\cellcolor[HTML]{D2D2E7}} \color[HTML]{000000} -0.02 & {\cellcolor[HTML]{6BA5CD}} \color[HTML]{F1F1F1} 0.09 & {\cellcolor[HTML]{A7BDDB}} \color[HTML]{000000} 0.04 & {\cellcolor[HTML]{2685BB}} \color[HTML]{F1F1F1} 0.06 & {\cellcolor[HTML]{E6E2EF}} \color[HTML]{000000} -0.01 & {\cellcolor[HTML]{DDDBEC}} \color[HTML]{000000} -0.03 & {\cellcolor[HTML]{9FBAD9}} \color[HTML]{000000} 0.07 & {\cellcolor[HTML]{B8C6E0}} \color[HTML]{000000} 0.05 \\
word\_ngram\_src\_overlap & {\cellcolor[HTML]{0872B1}} \color[HTML]{F1F1F1} 0.18 & {\cellcolor[HTML]{056DAC}} \color[HTML]{F1F1F1} 0.09 & {\cellcolor[HTML]{B7C5DF}} \color[HTML]{000000} 0.08 & {\cellcolor[HTML]{ABBFDC}} \color[HTML]{000000} 0.06 & {\cellcolor[HTML]{A9BFDC}} \color[HTML]{000000} 0.05 & {\cellcolor[HTML]{589EC8}} \color[HTML]{F1F1F1} 0.06 & {\cellcolor[HTML]{71A8CE}} \color[HTML]{F1F1F1} 0.09 & {\cellcolor[HTML]{4295C3}} \color[HTML]{F1F1F1} 0.14 & {\cellcolor[HTML]{056CAA}} \color[HTML]{F1F1F1} 0.18 & {\cellcolor[HTML]{3991C1}} \color[HTML]{F1F1F1} 0.15 & {\cellcolor[HTML]{3D93C2}} \color[HTML]{F1F1F1} 0.03 & {\cellcolor[HTML]{E7E3F0}} \color[HTML]{000000} -0.01 & {\cellcolor[HTML]{BDC8E1}} \color[HTML]{000000} 0.05 & {\cellcolor[HTML]{5A9EC9}} \color[HTML]{F1F1F1} 0.15 & {\cellcolor[HTML]{A5BDDB}} \color[HTML]{000000} 0.08 \\
word\_ngram\_repetition\_rate (REP-3) & {\cellcolor[HTML]{FFF7FB}} \color[HTML]{000000} -0.59 & {\cellcolor[HTML]{4A98C5}} \color[HTML]{F1F1F1} 0.05 & {\cellcolor[HTML]{FFF7FB}} \color[HTML]{000000} -0.13 & {\cellcolor[HTML]{FFF7FB}} \color[HTML]{000000} -0.19 & {\cellcolor[HTML]{FFF7FB}} \color[HTML]{000000} -0.23 & {\cellcolor[HTML]{FFF7FB}} \color[HTML]{000000} -0.31 & {\cellcolor[HTML]{FFF7FB}} \color[HTML]{000000} -0.08 & {\cellcolor[HTML]{FFF7FB}} \color[HTML]{000000} -0.13 & {\cellcolor[HTML]{E5E1EF}} \color[HTML]{000000} -0.05 & {\cellcolor[HTML]{FFF7FB}} \color[HTML]{000000} -0.11 & {\cellcolor[HTML]{FFF7FB}} \color[HTML]{000000} -0.27 & {\cellcolor[HTML]{FFF7FB}} \color[HTML]{000000} -0.05 & {\cellcolor[HTML]{D9D8EA}} \color[HTML]{000000} -0.02 & {\cellcolor[HTML]{B9C6E0}} \color[HTML]{000000} 0.04 & {\cellcolor[HTML]{D8D7E9}} \color[HTML]{000000} -0.01 \\
src\_rouge\_recall & {\cellcolor[HTML]{3790C0}} \color[HTML]{F1F1F1} 0.06 & {\cellcolor[HTML]{0569A5}} \color[HTML]{F1F1F1} 0.09 & {\cellcolor[HTML]{B7C5DF}} \color[HTML]{000000} 0.08 & {\cellcolor[HTML]{ACC0DD}} \color[HTML]{000000} 0.06 & {\cellcolor[HTML]{B1C2DE}} \color[HTML]{000000} 0.03 & {\cellcolor[HTML]{63A2CB}} \color[HTML]{F1F1F1} 0.04 & {\cellcolor[HTML]{A1BBDA}} \color[HTML]{000000} 0.05 & {\cellcolor[HTML]{60A1CA}} \color[HTML]{F1F1F1} 0.11 & {\cellcolor[HTML]{69A5CC}} \color[HTML]{F1F1F1} 0.09 & {\cellcolor[HTML]{65A3CB}} \color[HTML]{F1F1F1} 0.11 & {\cellcolor[HTML]{8CB3D5}} \color[HTML]{000000} -0.05 & {\cellcolor[HTML]{E4E1EF}} \color[HTML]{000000} -0.01 & {\cellcolor[HTML]{B4C4DF}} \color[HTML]{000000} 0.07 & {\cellcolor[HTML]{69A5CC}} \color[HTML]{F1F1F1} 0.13 & {\cellcolor[HTML]{A8BEDC}} \color[HTML]{000000} 0.08 \\
src\_rouge\_f1 & {\cellcolor[HTML]{348EBF}} \color[HTML]{F1F1F1} 0.07 & {\cellcolor[HTML]{05659F}} \color[HTML]{F1F1F1} 0.10 & {\cellcolor[HTML]{B0C2DE}} \color[HTML]{000000} 0.09 & {\cellcolor[HTML]{A5BDDB}} \color[HTML]{000000} 0.07 & {\cellcolor[HTML]{ACC0DD}} \color[HTML]{000000} 0.04 & {\cellcolor[HTML]{62A2CB}} \color[HTML]{F1F1F1} 0.04 & {\cellcolor[HTML]{A1BBDA}} \color[HTML]{000000} 0.05 & {\cellcolor[HTML]{60A1CA}} \color[HTML]{F1F1F1} 0.11 & {\cellcolor[HTML]{6DA6CD}} \color[HTML]{F1F1F1} 0.09 & {\cellcolor[HTML]{65A3CB}} \color[HTML]{F1F1F1} 0.11 & {\cellcolor[HTML]{8FB4D6}} \color[HTML]{000000} -0.05 & {\cellcolor[HTML]{E5E1EF}} \color[HTML]{000000} -0.01 & {\cellcolor[HTML]{B3C3DE}} \color[HTML]{000000} 0.07 & {\cellcolor[HTML]{69A5CC}} \color[HTML]{F1F1F1} 0.13 & {\cellcolor[HTML]{A8BEDC}} \color[HTML]{000000} 0.08 \\
src\_rouge\_precision & {\cellcolor[HTML]{2987BC}} \color[HTML]{F1F1F1} 0.10 & {\cellcolor[HTML]{2081B9}} \color[HTML]{F1F1F1} 0.07 & {\cellcolor[HTML]{6FA7CE}} \color[HTML]{F1F1F1} 0.20 & {\cellcolor[HTML]{89B1D4}} \color[HTML]{000000} 0.12 & {\cellcolor[HTML]{91B5D6}} \color[HTML]{000000} 0.09 & {\cellcolor[HTML]{5A9EC9}} \color[HTML]{F1F1F1} 0.05 & {\cellcolor[HTML]{B3C3DE}} \color[HTML]{000000} 0.03 & {\cellcolor[HTML]{78ABD0}} \color[HTML]{F1F1F1} 0.09 & {\cellcolor[HTML]{DEDCEC}} \color[HTML]{000000} -0.03 & {\cellcolor[HTML]{A1BBDA}} \color[HTML]{000000} 0.05 & {\cellcolor[HTML]{ACC0DD}} \color[HTML]{000000} -0.09 & {\cellcolor[HTML]{E0DEED}} \color[HTML]{000000} -0.00 & {\cellcolor[HTML]{A9BFDC}} \color[HTML]{000000} 0.09 & {\cellcolor[HTML]{A1BBDA}} \color[HTML]{000000} 0.07 & {\cellcolor[HTML]{B4C4DF}} \color[HTML]{000000} 0.06 \\
tgt\_fluency (CoLA) & {\cellcolor[HTML]{328DBF}} \color[HTML]{F1F1F1} 0.07 & {\cellcolor[HTML]{056BA7}} \color[HTML]{F1F1F1} 0.09 & {\cellcolor[HTML]{C4CBE3}} \color[HTML]{000000} 0.06 & {\cellcolor[HTML]{A2BCDA}} \color[HTML]{000000} 0.08 & {\cellcolor[HTML]{93B5D6}} \color[HTML]{000000} 0.09 & {\cellcolor[HTML]{023858}} \color[HTML]{F1F1F1} 0.35 & {\cellcolor[HTML]{023858}} \color[HTML]{F1F1F1} 0.25 & {\cellcolor[HTML]{69A5CC}} \color[HTML]{F1F1F1} 0.10 & {\cellcolor[HTML]{69A5CC}} \color[HTML]{F1F1F1} 0.09 & {\cellcolor[HTML]{308CBE}} \color[HTML]{F1F1F1} 0.16 & {\cellcolor[HTML]{023858}} \color[HTML]{F1F1F1} 0.23 & {\cellcolor[HTML]{034871}} \color[HTML]{F1F1F1} 0.20 & {\cellcolor[HTML]{79ABD0}} \color[HTML]{F1F1F1} 0.18 & {\cellcolor[HTML]{75A9CF}} \color[HTML]{F1F1F1} 0.12 & {\cellcolor[HTML]{7EADD1}} \color[HTML]{F1F1F1} 0.14 \\
src\_fluency\_diff & {\cellcolor[HTML]{3991C1}} \color[HTML]{F1F1F1} 0.06 & {\cellcolor[HTML]{167BB6}} \color[HTML]{F1F1F1} 0.08 & {\cellcolor[HTML]{DAD9EA}} \color[HTML]{000000} 0.00 & {\cellcolor[HTML]{CED0E6}} \color[HTML]{000000} -0.01 & {\cellcolor[HTML]{B9C6E0}} \color[HTML]{000000} 0.01 & {\cellcolor[HTML]{034D79}} \color[HTML]{F1F1F1} 0.29 & {\cellcolor[HTML]{056BA7}} \color[HTML]{F1F1F1} 0.18 & {\cellcolor[HTML]{96B6D7}} \color[HTML]{000000} 0.05 & {\cellcolor[HTML]{8EB3D5}} \color[HTML]{000000} 0.06 & {\cellcolor[HTML]{3F93C2}} \color[HTML]{F1F1F1} 0.14 & {\cellcolor[HTML]{034F7D}} \color[HTML]{F1F1F1} 0.19 & {\cellcolor[HTML]{023858}} \color[HTML]{F1F1F1} 0.21 & {\cellcolor[HTML]{A1BBDA}} \color[HTML]{000000} 0.10 & {\cellcolor[HTML]{A9BFDC}} \color[HTML]{000000} 0.06 & {\cellcolor[HTML]{A5BDDB}} \color[HTML]{000000} 0.08 \\
mean\_self\_sonar\_sim & {\cellcolor[HTML]{FDF5FA}} \color[HTML]{000000} -0.57 & {\cellcolor[HTML]{056AA6}} \color[HTML]{F1F1F1} 0.09 & {\cellcolor[HTML]{FFF7FB}} \color[HTML]{000000} -0.13 & {\cellcolor[HTML]{F4EDF6}} \color[HTML]{000000} -0.14 & {\cellcolor[HTML]{F2ECF5}} \color[HTML]{000000} -0.16 & {\cellcolor[HTML]{76AAD0}} \color[HTML]{F1F1F1} 0.02 & {\cellcolor[HTML]{B7C5DF}} \color[HTML]{000000} 0.03 & {\cellcolor[HTML]{D3D4E7}} \color[HTML]{000000} -0.02 & {\cellcolor[HTML]{DAD9EA}} \color[HTML]{000000} -0.03 & {\cellcolor[HTML]{D6D6E9}} \color[HTML]{000000} -0.02 & {\cellcolor[HTML]{2D8ABD}} \color[HTML]{F1F1F1} 0.05 & {\cellcolor[HTML]{CDD0E5}} \color[HTML]{000000} 0.02 & {\cellcolor[HTML]{D2D3E7}} \color[HTML]{000000} 0.00 & {\cellcolor[HTML]{ADC1DD}} \color[HTML]{000000} 0.06 & {\cellcolor[HTML]{D9D8EA}} \color[HTML]{000000} -0.01 \\
mean\_src\_sonar\_sim & {\cellcolor[HTML]{81AED2}} \color[HTML]{F1F1F1} -0.10 & {\cellcolor[HTML]{3790C0}} \color[HTML]{F1F1F1} 0.06 & {\cellcolor[HTML]{88B1D4}} \color[HTML]{000000} 0.16 & {\cellcolor[HTML]{B8C6E0}} \color[HTML]{000000} 0.03 & {\cellcolor[HTML]{B7C5DF}} \color[HTML]{000000} 0.02 & {\cellcolor[HTML]{79ABD0}} \color[HTML]{F1F1F1} 0.01 & {\cellcolor[HTML]{86B0D3}} \color[HTML]{000000} 0.07 & {\cellcolor[HTML]{81AED2}} \color[HTML]{F1F1F1} 0.08 & {\cellcolor[HTML]{C4CBE3}} \color[HTML]{000000} 0.00 & {\cellcolor[HTML]{C6CCE3}} \color[HTML]{000000} 0.01 & {\cellcolor[HTML]{8CB3D5}} \color[HTML]{000000} -0.05 & {\cellcolor[HTML]{D9D8EA}} \color[HTML]{000000} 0.01 & {\cellcolor[HTML]{A9BFDC}} \color[HTML]{000000} 0.09 & {\cellcolor[HTML]{B7C5DF}} \color[HTML]{000000} 0.04 & {\cellcolor[HTML]{AFC1DD}} \color[HTML]{000000} 0.07 \\
mean\_src\_coverage\_sonar\_sim & {\cellcolor[HTML]{2A88BC}} \color[HTML]{F1F1F1} 0.09 & {\cellcolor[HTML]{0D75B3}} \color[HTML]{F1F1F1} 0.08 & {\cellcolor[HTML]{AFC1DD}} \color[HTML]{000000} 0.10 & {\cellcolor[HTML]{93B5D6}} \color[HTML]{000000} 0.10 & {\cellcolor[HTML]{9AB8D8}} \color[HTML]{000000} 0.07 & {\cellcolor[HTML]{89B1D4}} \color[HTML]{000000} -0.01 & {\cellcolor[HTML]{71A8CE}} \color[HTML]{F1F1F1} 0.09 & {\cellcolor[HTML]{5A9EC9}} \color[HTML]{F1F1F1} 0.11 & {\cellcolor[HTML]{6DA6CD}} \color[HTML]{F1F1F1} 0.09 & {\cellcolor[HTML]{97B7D7}} \color[HTML]{000000} 0.06 & {\cellcolor[HTML]{88B1D4}} \color[HTML]{000000} -0.04 & {\cellcolor[HTML]{D5D5E8}} \color[HTML]{000000} 0.01 & {\cellcolor[HTML]{A4BCDA}} \color[HTML]{000000} 0.10 & {\cellcolor[HTML]{A8BEDC}} \color[HTML]{000000} 0.06 & {\cellcolor[HTML]{9EBAD9}} \color[HTML]{000000} 0.09 \\
mean\_monotonic\_src\_sonar\_sim & {\cellcolor[HTML]{2A88BC}} \color[HTML]{F1F1F1} 0.10 & {\cellcolor[HTML]{79ABD0}} \color[HTML]{F1F1F1} 0.03 & {\cellcolor[HTML]{67A4CC}} \color[HTML]{F1F1F1} 0.21 & {\cellcolor[HTML]{84B0D3}} \color[HTML]{F1F1F1} 0.13 & {\cellcolor[HTML]{8BB2D4}} \color[HTML]{000000} 0.11 & {\cellcolor[HTML]{7BACD1}} \color[HTML]{F1F1F1} 0.01 & {\cellcolor[HTML]{89B1D4}} \color[HTML]{000000} 0.07 & {\cellcolor[HTML]{7DACD1}} \color[HTML]{F1F1F1} 0.08 & {\cellcolor[HTML]{C4CBE3}} \color[HTML]{000000} 0.00 & {\cellcolor[HTML]{C2CBE2}} \color[HTML]{000000} 0.01 & {\cellcolor[HTML]{8CB3D5}} \color[HTML]{000000} -0.05 & {\cellcolor[HTML]{D8D7E9}} \color[HTML]{000000} 0.01 & {\cellcolor[HTML]{ABBFDC}} \color[HTML]{000000} 0.09 & {\cellcolor[HTML]{B8C6E0}} \color[HTML]{000000} 0.04 & {\cellcolor[HTML]{AFC1DD}} \color[HTML]{000000} 0.07 \\
sonar\_sim\_monotonicity & {\cellcolor[HTML]{034D79}} \color[HTML]{F1F1F1} 0.37 & {\cellcolor[HTML]{FFF7FB}} \color[HTML]{000000} -0.07 & {\cellcolor[HTML]{8BB2D4}} \color[HTML]{000000} 0.16 & {\cellcolor[HTML]{84B0D3}} \color[HTML]{F1F1F1} 0.13 & {\cellcolor[HTML]{8BB2D4}} \color[HTML]{000000} 0.10 & {\cellcolor[HTML]{96B6D7}} \color[HTML]{000000} -0.04 & {\cellcolor[HTML]{D5D5E8}} \color[HTML]{000000} -0.00 & {\cellcolor[HTML]{7BACD1}} \color[HTML]{F1F1F1} 0.08 & {\cellcolor[HTML]{9AB8D8}} \color[HTML]{000000} 0.05 & {\cellcolor[HTML]{9AB8D8}} \color[HTML]{000000} 0.06 & {\cellcolor[HTML]{73A9CF}} \color[HTML]{F1F1F1} -0.02 & {\cellcolor[HTML]{9EBAD9}} \color[HTML]{000000} 0.05 & {\cellcolor[HTML]{DFDDEC}} \color[HTML]{000000} -0.04 & {\cellcolor[HTML]{F2ECF5}} \color[HTML]{000000} -0.07 & {\cellcolor[HTML]{C9CEE4}} \color[HTML]{000000} 0.02 \\
p\_entail\_full & {\cellcolor[HTML]{5EA0CA}} \color[HTML]{F1F1F1} -0.02 & {\cellcolor[HTML]{89B1D4}} \color[HTML]{000000} 0.02 & {\cellcolor[HTML]{056AA6}} \color[HTML]{F1F1F1} 0.38 & {\cellcolor[HTML]{6DA6CD}} \color[HTML]{F1F1F1} 0.17 & {\cellcolor[HTML]{4E9AC6}} \color[HTML]{F1F1F1} 0.20 & {\cellcolor[HTML]{96B6D7}} \color[HTML]{000000} -0.04 & {\cellcolor[HTML]{D3D4E7}} \color[HTML]{000000} 0.00 & {\cellcolor[HTML]{056DAC}} \color[HTML]{F1F1F1} 0.21 & {\cellcolor[HTML]{71A8CE}} \color[HTML]{F1F1F1} 0.08 & {\cellcolor[HTML]{4697C4}} \color[HTML]{F1F1F1} 0.14 & {\cellcolor[HTML]{78ABD0}} \color[HTML]{F1F1F1} -0.02 & {\cellcolor[HTML]{E0DDED}} \color[HTML]{000000} -0.00 & {\cellcolor[HTML]{056AA6}} \color[HTML]{F1F1F1} 0.39 & {\cellcolor[HTML]{0569A4}} \color[HTML]{F1F1F1} 0.26 & {\cellcolor[HTML]{056BA7}} \color[HTML]{F1F1F1} 0.31 \\
p\_noncontradict\_full & {\cellcolor[HTML]{60A1CA}} \color[HTML]{F1F1F1} -0.02 & {\cellcolor[HTML]{E2DFEE}} \color[HTML]{000000} -0.03 & {\cellcolor[HTML]{6BA5CD}} \color[HTML]{F1F1F1} 0.20 & {\cellcolor[HTML]{B3C3DE}} \color[HTML]{000000} 0.04 & {\cellcolor[HTML]{A4BCDA}} \color[HTML]{000000} 0.06 & {\cellcolor[HTML]{AFC1DD}} \color[HTML]{000000} -0.08 & {\cellcolor[HTML]{F5EEF6}} \color[HTML]{000000} -0.05 & {\cellcolor[HTML]{84B0D3}} \color[HTML]{F1F1F1} 0.07 & {\cellcolor[HTML]{D9D8EA}} \color[HTML]{000000} -0.03 & {\cellcolor[HTML]{CDD0E5}} \color[HTML]{000000} -0.00 & {\cellcolor[HTML]{529BC7}} \color[HTML]{F1F1F1} 0.01 & {\cellcolor[HTML]{F8F1F8}} \color[HTML]{000000} -0.04 & {\cellcolor[HTML]{4697C4}} \color[HTML]{F1F1F1} 0.25 & {\cellcolor[HTML]{3B92C1}} \color[HTML]{F1F1F1} 0.18 & {\cellcolor[HTML]{4697C4}} \color[HTML]{F1F1F1} 0.20 \\
smollm\_reconstruct\_loss & {\cellcolor[HTML]{7BACD1}} \color[HTML]{F1F1F1} -0.09 & {\cellcolor[HTML]{F4EDF6}} \color[HTML]{000000} -0.05 & {\cellcolor[HTML]{FBF3F9}} \color[HTML]{000000} -0.11 & {\cellcolor[HTML]{D3D4E7}} \color[HTML]{000000} -0.03 & {\cellcolor[HTML]{CDD0E5}} \color[HTML]{000000} -0.03 & {\cellcolor[HTML]{84B0D3}} \color[HTML]{F1F1F1} -0.01 & {\cellcolor[HTML]{D9D8EA}} \color[HTML]{000000} -0.01 & {\cellcolor[HTML]{E6E2EF}} \color[HTML]{000000} -0.06 & {\cellcolor[HTML]{FFF7FB}} \color[HTML]{000000} -0.11 & {\cellcolor[HTML]{F0EAF4}} \color[HTML]{000000} -0.07 & {\cellcolor[HTML]{308CBE}} \color[HTML]{F1F1F1} 0.05 & {\cellcolor[HTML]{D2D2E7}} \color[HTML]{000000} 0.02 & {\cellcolor[HTML]{FFF7FB}} \color[HTML]{000000} -0.17 & {\cellcolor[HTML]{FFF7FB}} \color[HTML]{000000} -0.11 & {\cellcolor[HTML]{FFF7FB}} \color[HTML]{000000} -0.13 \\
seahorse\_q4 (SH-4) & {\cellcolor[HTML]{0569A4}} \color[HTML]{F1F1F1} 0.24 & {\cellcolor[HTML]{83AFD3}} \color[HTML]{F1F1F1} 0.03 & {\cellcolor[HTML]{023858}} \color[HTML]{F1F1F1} 0.52 & {\cellcolor[HTML]{0568A3}} \color[HTML]{F1F1F1} 0.37 & {\cellcolor[HTML]{045F95}} \color[HTML]{F1F1F1} 0.41 & {\cellcolor[HTML]{056DAC}} \color[HTML]{F1F1F1} 0.19 & {\cellcolor[HTML]{81AED2}} \color[HTML]{F1F1F1} 0.08 & {\cellcolor[HTML]{023858}} \color[HTML]{F1F1F1} 0.31 & {\cellcolor[HTML]{1077B4}} \color[HTML]{F1F1F1} 0.17 & {\cellcolor[HTML]{03517E}} \color[HTML]{F1F1F1} 0.27 & {\cellcolor[HTML]{04649E}} \color[HTML]{F1F1F1} 0.14 & {\cellcolor[HTML]{94B6D7}} \color[HTML]{000000} 0.06 & {\cellcolor[HTML]{023858}} \color[HTML]{F1F1F1} 0.54 & {\cellcolor[HTML]{023858}} \color[HTML]{F1F1F1} 0.35 & {\cellcolor[HTML]{023858}} \color[HTML]{F1F1F1} 0.43 \\
seahorse\_q5 (SH-5) & {\cellcolor[HTML]{023E62}} \color[HTML]{F1F1F1} 0.43 & {\cellcolor[HTML]{7BACD1}} \color[HTML]{F1F1F1} 0.03 & {\cellcolor[HTML]{1379B5}} \color[HTML]{F1F1F1} 0.33 & {\cellcolor[HTML]{023858}} \color[HTML]{F1F1F1} 0.51 & {\cellcolor[HTML]{034C78}} \color[HTML]{F1F1F1} 0.47 & {\cellcolor[HTML]{056BA9}} \color[HTML]{F1F1F1} 0.20 & {\cellcolor[HTML]{4697C4}} \color[HTML]{F1F1F1} 0.12 & {\cellcolor[HTML]{045E94}} \color[HTML]{F1F1F1} 0.25 & {\cellcolor[HTML]{023858}} \color[HTML]{F1F1F1} 0.27 & {\cellcolor[HTML]{03466E}} \color[HTML]{F1F1F1} 0.29 & {\cellcolor[HTML]{045C90}} \color[HTML]{F1F1F1} 0.16 & {\cellcolor[HTML]{1379B5}} \color[HTML]{F1F1F1} 0.14 & {\cellcolor[HTML]{0570B0}} \color[HTML]{F1F1F1} 0.36 & {\cellcolor[HTML]{045687}} \color[HTML]{F1F1F1} 0.30 & {\cellcolor[HTML]{0569A5}} \color[HTML]{F1F1F1} 0.31 \\
seahorse\_q6 & {\cellcolor[HTML]{023858}} \color[HTML]{F1F1F1} 0.45 & {\cellcolor[HTML]{A1BBDA}} \color[HTML]{000000} 0.01 & {\cellcolor[HTML]{045E94}} \color[HTML]{F1F1F1} 0.42 & {\cellcolor[HTML]{023858}} \color[HTML]{F1F1F1} 0.51 & {\cellcolor[HTML]{023858}} \color[HTML]{F1F1F1} 0.52 & {\cellcolor[HTML]{045B8E}} \color[HTML]{F1F1F1} 0.26 & {\cellcolor[HTML]{60A1CA}} \color[HTML]{F1F1F1} 0.10 & {\cellcolor[HTML]{023B5D}} \color[HTML]{F1F1F1} 0.31 & {\cellcolor[HTML]{045B8E}} \color[HTML]{F1F1F1} 0.22 & {\cellcolor[HTML]{023858}} \color[HTML]{F1F1F1} 0.31 & {\cellcolor[HTML]{03466E}} \color[HTML]{F1F1F1} 0.20 & {\cellcolor[HTML]{529BC7}} \color[HTML]{F1F1F1} 0.10 & {\cellcolor[HTML]{045382}} \color[HTML]{F1F1F1} 0.47 & {\cellcolor[HTML]{023A5B}} \color[HTML]{F1F1F1} 0.35 & {\cellcolor[HTML]{034267}} \color[HTML]{F1F1F1} 0.41 \\
\bottomrule
\end{tabular}
\caption{\label{tab:metrics_early_corrs} Spearman correlations of candidate automatic metrics with human judgments on the English subset of SEAHORSE \citep{clark-etal-2023-seahorse}, grouped by subsets of the data.}
\end{table*}

\section{Summarization Guidelines}
\label{human:summguide}
\paragraph{Annotator proficiency requirements}
\begin{itemize}
\item Native speaker of English
\item Editor / writer / domain expert
\end{itemize}

\paragraph{Task}
You will receive document(s) that are approximately 5,000 words or longer from the following domains:

\begin{itemize}
\item Political (GovReports, MultiUN)
\item News (Seahorse)
\item Wikipedia (Wikipedia)
\item Scientific/Technical (FacetSum) 
\item Literature (BookSum, SQuality)
\item Conversational (Summscreen)
\item Legal (LexGlue, JRCAcquis)
\end{itemize}

The documents will contain sections/chapters. You will need to summarize them retaining the section alignment. For certain domains, there will be additional guidance in the form of special guidelines (legal, medical etc.)

You will need to create 3 summaries: 
\begin{itemize}
\item Summary 1: around 20\% of the source text (~1,000 words if total length is ~5000) 
\item Summary 2: around 10\% of the source text (~500 words)
\item Summary 3: around 5\% of the source text (~250 words)
\end{itemize}

After finishing summarizing, you will need to write a minimum of 15 questions with corresponding answers (QA) per each 5000 words. 

\paragraph{Requirements for Summarization}

Here’s more information on what that means and how to summarize:

Please read the provided document in its entirety. Consider making notes of the main core ideas while you read. After you have finished reading, please write a short summary of the source text. The summary should:
\begin{itemize}
\item Be much shorter than the source text. Please see the information about the length above.
\item Convey ALL the main core ideas and information of the source document.
\item Have a structure of a standalone cohesive text.
\item Follow the flow of the section/paragraph structure of the source text, try to summarize it  from top to bottom.
\item Each paragraph in each summary should be marked with a number of the source text section/chapter showing where this information is from.
\end{itemize}

Here’s a checklist which can help you with the task:
\begin{itemize}
\item \textbf{Understand the Main Idea}: Read the entire text to grasp the overall theme and the author's intent. Identify the main idea of the text.
\item \textbf{Highlight Key Points}: Mark or note down the essential points and arguments. 
\item \textbf{Eliminate Redundancies}: Remove any repetitive information or examples that do not add value to the understanding of the main idea.
\item \textbf{Use Your Own Words}: Paraphrase the key points in your own words instead of copying verbatim. This helps ensure the summary is concise. 
\item \textbf{Keep It Objective}: Focus on the information presented in the text without inserting personal opinions or interpretations.
\item \textbf{Structure the Summary}: Organize the ideas logically, maintaining the flow of the original text.
\item \textbf{Be Concise}: Aim for clarity and brevity; Use simple and direct language to convey the points. 
\item \textbf{Review and Revise}: Compare the summary to the original text to ensure accuracy and completeness. Edit for coherence, transitions, and readability.
\end{itemize}

\paragraph{Additional Guidelines for Conversational Text}
You may be assigned to work on conversational type of text, such as meeting transcripts, screenplays, and novels. Since the text structure is quite different from documents, this is an additional guideline to help you working on conversational text:
\begin{itemize}
\item Skim through the whole document: Try to get a rough idea of the whole plot 
\item Identify the characters and main core ideas: Identify the main characters and focus on their interaction
\item Omit the trivial details: there maybe side plot or supporting characters in the source text, carefully decide if it is related to the main core ideas (plot)
\item Group and summarize with respect to main core ideas: There could be plot twists or related hints in the source documents. Remember the summary should be clear and straightforward, the plot outline \item Should be clear in the summary without referencing to the source document
\end{itemize}

\paragraph{Requirements for Question and Answer sets}
\begin{itemize}
\item The questions need to be abstractive not extractive: this means they need to be directed at the ideas in the text, not words and sentences as such.
\item The questions should be open-ended, not Yes/No questions
\item The correct answers should cover the main points in the source text: the questions should roughly correspond to paragraphs/sections in the text
\item Thus, the correct answer should cover points reflected in the summary (for your convenience, you can refer to your longest summary, but please mind your shorter summary also need to be able to answer at least some of the questions)
\item The answers should not be short (30 words or more)
\item The correct answer should be found namely IN THE SUMMARY and not able to be just pulled from general knowledge
\item If possible, refrain from factual questions, but try composing questions for reasoning, such as WHY- questions
\item You are encouraged to combine information from different sections together
\item Avoid only asking questions about the beginning and the end of the section, use the information in the middle as well
\item If possible, the questions should not have several answer possibilities.
\end{itemize}

\section{Human Summary Sample}
This is one sample of 3 human written summaries and QA set of the document 4586 from GovReport. 

\subsection{Source Document Excerpt} 
\textit{[This is the first and last paragraphs of the source document. The whole document is 5411 words long.]}

This report examines technological innovation in payment systems generally and particular policy issues as a result of retail (i.e., point of sale) payment innovation. The report also discusses wholesale payment, clearing, and settlement systems that send payment messages between banks and transfer funds, including the "real-time payments" service being introduced by the Federal Reserve. This report includes an Appendix that describes interbank payment, clearing, and settlement systems related to U.S. payments.

...

To address systemic risk concerns, a private RTP system could be designated as a systemically important Financial Market Utility (FMU) under Title VIII of the Dodd-Frank Act ( P.L. 111-203 ). The Dodd-Frank Act allows the Financial Stability Oversight Council , a council of financial regulators led by the Treasury Secretary, to designate a payment, clearing, or settlement system as systemically important on the grounds that "the failure of or a disruption to the functioning of the FMU could create or increase the risk of significant liquidity or credit problems spreading among financial institutions or markets and thereby threaten the stability of the U.S. financial system." FMUs, currently including the Clearing House Interbank Payments System, are subject to heightened regulation, and the Fed has supervisory and enforcement powers to ensure those standards are met. Policymakers could consider whether systemic risk concerns are better addressed through Fed operation of payment and settlement systems or Fed regulation of private systems.

\subsection{Long, Medium and Short Summaries} 

Each paragraph of the summaries is paired with the source paragraph id (e.g. p1, p2, etc) to indicate the information source.
\paragraph{Long Summary (20\%)}

To the average consumer, swiping their credit card seems simple, because the complex the infrastructure involved in is 'hidden'. These deceptively "simple" electronic payments are comprised of three main steps. First, the sender makes the payment through an online payment service or an app, which instructs the sender's bank to make the payment to the recipient. Second, the bank sends a payment message to the recipient's bank through a payment system or clearing service. Finally, the payment is completed (settled) when the funds are received by the recipient. (p2)

Some of the bank-to-bank (ACH) payment, clearing, and settlement (PCS) systems are operated by the Federal Reserve, and others by private-sector organizations. Recently, the use of electronic payment methods (credit card, debit card, and ACH) has grown, while the use cash and check payments has declined. Electronic payments have been made easier and more convenient with digital wallets and payment apps like Venmo, Cash App, and Zelle - all of which require users to link a bank account, credit card, or debit card. (p4, p5, p6, p7)

There are concerns about whether current regulations are equipped to handle electronic payments. If not, this poses potential risks to cybersecurity, data privacy, industry competition, and consumer access and protection. Current payment regulations depend, in part, on if the service is provided by a bank, who have many strict regulatory requirements. As such, Nonbank payment systems are not subject to existing regulatory enforcement and can only be supervised - as money transmitters at state level and money service businesses at federal level. (p8, p9)

Electronic payment regulatory concerns could be addressed by including nonbank payment companies into the bank regulatory regime. One way could be via the Office of the Comptroller of the Currency (OCC) special purpose national bank charter. And another, through a state-level industrial loan company (ILC) charter with the Federal Deposit Insurance Corporation (FDIC). Both methods could provide nonbank firms access to the Fed wholesale payment systems, which could be advantageous. However, some state regulators have filed lawsuits to block nonbank companies access to these charters, arguing that it allows companies to circumvent state consumer protections. So far, no companies have applied for an OCC charter, likely due to the legal uncertainty surrounding it. (p11, p12)

The main argument against nonbank payment companies filing ILC charters is that it would allow them to own banks - and the FDIC has not approved deposit insurance for a new ILC since 2006. Opponents argue that allowing a company to own a bank could expose the US economy to risks like imprudent underwriting. Proponents assert that these concerns are exaggerated, noting that several other countries allow similar arrangement with no ill effects. So far, Square is the only company with a pending application and two other companies have withdrawn their applications. (p13)

New technology reduces some risks related to payments but creates new ones. The risk of having one's wallet stolen is reduced, but payment information is subject to more sophisticated risks such as malware attacks. Furthermore, storing payment information on a variety of websites, apps, and devices creates more opportunities for hackers. After recent security breaches which allowed user information to be stolen, several solutions have been proposed. For example, a federal breach notification law could be enacted, to create federal cybersecurity standards or to increase penalties for companies with inadequate security measures. (p15, p16, p17)

Payment systems need to collect detailed information about customer transactions in order to function properly. This data can be used by companies to target ads. Scammers can also use this information for fraudulent purposes. The constantly increasing use of Electronic payments has led to questions about how user data is used and whether consumers are sufficiently informed and given enough control about how their data is used. (p18)

There are some consumer benefits to storing consumer data. It can help them track payments and budget more easily by importing to budgeting apps. They can also share financial information with banks more easily when applying for loans. But, given the benefits and the risks, the question remains: how much access should companies have to individuals' information? (p19)

Privacy policies are another area of concern with respect to consumer protection and electronic payments. According to the Bureau of Consumer Financial Protection (CFPB), it is difficult to provide disclosures that are clear and easy to understand, partly due to the small screens on phones. Clearer privacy policies and allowing consumers more control over how their data is used could help. (p20)

The Electronic Fund Transfer Act, Regulation E implemented by the CFPB, is the most relevant law aimed at protecting consumers who are making electronic payments. It mandates consumer disclosures, limits consumer liability for unauthorized payments, and maintains procedures for resolving errors. Further regulations are being considered. (p22)

Consumers could also be protected through financial education, especially for more at-risk older and lower-income groups. This could include learning how to use new payment systems safely and how to protect against financial harm, as well as knowing how to get help if something goes wrong. (p24)

Payment system innovations may affect consumers differently based on income. Consumers who mainly pay with cash, don't have bank accounts, or don't have internet or mobile access won't be able to benefit. Neither will those who are not comfortable using new technology. (p25)

However, surveys reveal that 83\% of underbanked, and 50\% of unbanked, consumers have smart phone access. So, as costs of these payment services decline, some marginalized groups could experience better access to the the financial system through access to digital currency channels via cash equivalents like pre-paid cards. But, the cost of internet and mobile data plans may limit access to faster payment systems, so this also needs to be considered. (p26, p27, p28)

Faster payment systems may also benefit low-income consumers by allowing them faster access to their paychecks and other fund transfers. But a potential drawback is that withdrawals from their accounts would occur more quickly as well. (p28)

In 2019, the Fed announced that it plans to create a wholesale real-time payment (RTP) system. (p32)

Originally, the Fed's primary function was to provide bank-to-bank check-clearing services. Private clearing houses were experiencing issues that led to the creation of the Fed. As payment methods have evolved, the Fed has begun providing other types of payment systems. It does this by linking the accounts that all banks keep at the Fed so that it can complete the transfers. The new system that the Fed is developing, called FedNow, would allow payments to occur in real time, rather than later in the day - or even the next day, as is the case currently. (p33, p35)

However, there are some concerns regarding implementation of FedNow. Many worry that it will undermine private sector development of similar systems. Others fear that failing to implement FedNow will lead to a monopoly of a private-sector company, to the detriment of consumers and smaller banks. (p42, p44)

\paragraph{Medium Summary (10\%)}

Electronic payments have three stages. First, the sender makes the payment through an online payment service or an app, which instructs the sender's bank to make the payment to the recipient. Second, the bank sends a payment message to the recipient's bank through a payment system or clearing service. Finally, the payment is completed (settled) when the funds are received by the recipient. (p2)

Some of the bank-to-bank (ACH) payment, clearing, and settlement (PCS) systems are operated by the Federal Reserve, and others by private-sector organizations. Recently, the use of electronic payment methods (credit card, debit card, and ACH) has grown, while the use cash and check payments has declined. Electronic payments have been made easier and more convenient with digital wallets and payment apps like Venmo, Cash App, and Zelle - all of which require users to link a bank account, credit card, or debit card.  (p4, p5, p6, p7)

There is concern about whether current regulations are equipped to handle these technological advances. If not, they could pose risks to cybersecurity, data privacy, industry competition, and consumer access and protection.  (p8)

One way to address these concerns is to add nonbank companies to the bank regulatory regime. Another is  via the Office of the Comptroller of the Currency (OCC) special purpose national bank charter. And another, through a state-level industrial loan company (ILC) charter with the Federal Deposit Insurance Corporation (FDIC). Both could provide nonbank firms access to the Fed wholesale payment systems, which could be advantageous. However, some state regulators have tried to block nonbank access to these charters, arguing that it allows companies to circumvent state consumer protections. So far, no companies have applied for an OCC charter, likely due to the legal uncertainty surrounding it.  (p11, p12)

The main argument against ILC charters is that it would allow companies to own banks. The FDIC has not approved deposit insurance for a new ILC since 2006. Opponents argue that allowing a retailer to own a bank could expose the US economy to risks such as imprudent underwriting. Proponents assert that these concerns are exaggerated and that several other countries allow similar arrangement with no ill effects. Currently, Square is the only company with a pending application.  (p13)

Privacy policies are another area of concern. According to the Bureau of Consumer Financial Protection (CFPB), it is difficult to provide disclosures that are clear and easy to understand, partly due to the small screens on phones. Clear privacy policies and allowing consumers more control over how their data is used could help.  (p20)

The Electronic Fund Transfer Act, Regulation E implemented by the CFPB, is the most relevant law aimed at protecting consumers who are making electronic payments. It mandates consumer disclosures, limits consumer liability for unauthorized payments, and maintains procedures for resolving errors.  (p22)

In 2019, the Fed announced that it plans to create a wholesale real-time payment (RTP) system.The proposed system, called FedNow, would allow payments to occur in real time, rather than later in the day - or even the next day, as is the case currently.  (p32, p33, p35)

There are some concerns regarding FedNow. Many worry that it will undermine private sector development of similar systems. Others fear that failing to implement FedNow will lead to a monopoly of a private-sector company, to the detriment of consumers and smaller banks.  (p42, p44)

\paragraph{Short Summaries (5\%)}

Electronic payments have three stages. First, the sender makes the payment through an online payment service or an app, which instructs the sender's bank to make the payment to the recipient. Second, the bank sends a payment message to the recipient's bank through a payment system or clearing service. Finally, the payment is completed (settled) when the funds are received by the recipient. (p2)

Some of the bank-to-bank payment, clearing, and settlement (PCS) systems are operated by the Federal Reserve, and others by private-sector organizations. Recently, the use of electronic payment methods has grown, while the use cash and check payments has declined.  (p4)

There is concern about whether current regulations are equipped to handle these technological advances. If not, they could pose risks to cybersecurity, data privacy, industry competition, and consumer access and protection.  (p8, p9)

A potential way to resolve concerns regarding regulation is to add certain nonbank payment companies into the bank regulatory regime. One way to accomplish this is through the OCC or Office of the Comptroller of the Currency special purpose national bank charter. The second is through a state-level industrial loan company (ILC) charter with the Federal Deposit Insurance Corporation (FDIC).  (p11, p12)

Privacy policies are another area of concern. According to the Bureau of Consumer Financial Protection (CFPB), it is difficult to provide disclosures that are clear and easy to understand, partly due to the small screens on phones. Clear privacy policies and allowing consumers more control over how their data is used could help.  (p20)

In 2019, the Fed announced that it plans to create a wholesale real-time payment (RTP) system. The proposed system, called FedNow, would allow payments to occur in real time, rather than later in the day - or even the next day as is the case currently.  (p32, p33, p35)

\subsection{Question and Answer Set}
\textbf{Question 1}: What are the three parts of a payment system?
\begin{itemize}
\item \textbf{Answer}: First, there is the sender or the person making the payment through an online payment service or an app, which instructs the sender's bank to make the payment to the recipient. Second, the bank sends a payment message to the recipient's bank through a payment system or clearing service. Finally, the payment is completed when the funds are transferred, or settled. 
\item (Information contained in 20\% Summary, 10\% Summary, 5\% Summary.)
\item (Source paragraph number: p2)
\end{itemize}
\textbf{Question 2}: Who operates bank-to-bank payment, clearing, and settlement systems?
\begin{itemize}
\item \textbf{Answer}: Some of these systems are operated by the Federal Reserve and some are operated by private-sector organizations.
\item (Information contained in 20\% Summary, 10\% Summary, 5\% Summary.)
\item (Source paragraph number: p4)
\end{itemize}
\textbf{Question 3}: What issues could there be if current regulations are not equipped to handle these payment system innovations?
\begin{itemize}
\item \textbf{Answer}: If regulations are inadequate, there could be issues related to cybersecurity, data privacy, industry competition, and consumer access and protection. 
\item (Information contained in 20\% Summary, 10\% Summary, 5\% Summary.)
\item (Source paragraph number: p8)
\end{itemize}
\textbf{Question 4}: What are two ways to bring nonbank companies into the bank regulatory regime?
\begin{itemize}
\item \textbf{Answer}: One way to accomplish this is through the OCC or Office of the Comptroller of the Currency special purpose national bank charter. The second is through a state-level industrial loan company (ILC) charter with the Federal Deposit Insurance Corporation (FDIC). 
\item (Information contained in 20\% Summary, 10\% Summary, 5\% Summary.)
\item (Source paragraph number: p11,12)
\end{itemize}
\textbf{Question 5}: According to the Bureau of Consumer Financial Protection, what are some of the difficulties with privacy policies?
\begin{itemize}
\item \textbf{Answer}: It is difficult to provide disclosures that are clear and easy to understand, partly due to the small screens on phones.
\item (Information contained in 20\% Summary, 10\% Summary, 5\% Summary.)
\item (Source paragraph number: p20)
\end{itemize}
\textbf{Question 6}: What is FedNow?
\begin{itemize}
\item \textbf{Answer}: In 2019, the Fed announced that it plans to create a wholesale real-time payment (RTP) system. The proposed system, called FedNow, would allow payments to occur in real time, rather than later in the day or even the next day as is the case currently. 
\item (Information contained in 20\% Summary, 10\% Summary, 5\% Summary.)
\item (Source paragraph number: p29)
\end{itemize}
\textbf{Question 7}: What are some reasons for the increase in electronic payments?
\begin{itemize}
\item \textbf{Answer}: Electronic payments have increased because of payment apps such as Venmo, Cash App, and Zelle make it convenient and easy for consumers to send payments. Digital wallets stored on phones are another reason for increased electronic payments due their ease of use and convenience. 
\item (Information contained in 20\% Summary, 10\% Summary.)
\item (Source paragraph number: p6)
\end{itemize}
\textbf{Question 8}: What is necessary in order for a consumer to be able to use electronic payment services?
\begin{itemize}
\item \textbf{Answer}: The consumer must have a debit card, credit card, or bank account linked to an electronic payment system. 
\item (Information contained in 20\% Summary, 10\% Summary.)
\item (Source paragraph number: p7)
\end{itemize}
\textbf{Question 9}: Why have state regulators filed lawsuits to block the OCC?
\begin{itemize}
\item \textbf{Answer}: Regulators feel that the OCC charter would allow companies to avoid state regulations that protect consumers. 
\item (Information contained in 20\% Summary, 10\% Summary.)
\item (Source paragraph number: p11, p12)
\end{itemize}
\textbf{Question 10}: What is the main argument against the ILC charter?
\begin{itemize}
\item \textbf{Answer}: The ILC would allow companies such as retailers to own banks. Opponents are concerned that this could lead to imprudent underwriting and could hurt the US economy by exposing it to risk.
\item (Information contained in 20\% Summary, 10\% Summary.)
\item (Source paragraph number: p13)
\end{itemize}
\textbf{Question 11}: What does the Electronic Funds Transfer Act Regulation E do?
\begin{itemize}
\item \textbf{Answer}: Regulation E mandates consumer disclosures, limits consumer liability for unauthorized payments, and maintains procedures for resolving errors. 
\item (Information contained in 20\% Summary, 10\% Summary.)
\item (Source paragraph number: p22)
\end{itemize}
\textbf{Question 12}: How could financial education help consumers use electronic payment systems safely?
\begin{itemize}
\item \textbf{Answer}: Consumers could be taught how to use new payment systems safely and how to protect against financial harm, as well as how to get help if something goes wrong. 
\item (Information contained in 20\% Summary.)
\item (Source paragraph number: p24)
\end{itemize}
\textbf{Question 13}: What are is an argument against the FedNow?
\begin{itemize}
\item \textbf{Answer}: Many worry that it will undermine private sector development of similar systems. 
\item (Information contained in 20\% Summary, 10\% Summary.)
\item (Source paragraph number: p42, p44)
\end{itemize}
\textbf{Question 14}: How can storing more consumer data benefit consumers?
\begin{itemize}
\item \textbf{Answer}: It can help consumers track payments and budget more easily using budgeting apps. They can also share financial information with banks more easily when applying for loans. 
\item (Information contained in 20\% Summary.)
\item (Source paragraph number: p19)
\end{itemize}
\textbf{Question 15}: How could faster payment systems affect low-income consumers?
\begin{itemize}
\item \textbf{Answer}: Faster payment systems may benefit low-income consumers by allowing them faster access to their paychecks and other fund transfers. But a potential drawback is that withdrawals from their accounts would occur more quickly as well. 
\item (Information contained in 20\% Summary.)
\item (Source paragraph number: p28)
\end{itemize}

\section{Summarization Human Evaluation Guidelines}
\label{app:humaneval:summ}

\paragraph{Annotator proficiency requirements}
All annotators must meet ALL of the following requirements:
\begin{itemize}
\item Native speaker of English AND
\item Language related degree holder or related professionals
\end{itemize}
\paragraph{Background Information}

What is a good summary? A good summary should meet the following criteria:
\begin{itemize}
\item \textbf{Conciseness}: The summary should only contain the most important information while maintaining readability. Trivial information should not be included, even in the longest summary. Additionally, the summary should be comprehensible on its own, without needing to refer to additional documentation.
\item \textbf{Coverage of main core ideas}: The summary should preserve the most important ideas, regardless of its length. In our task, summaries are created by gradually omitting less important information. Therefore, we expect that the core ideas will be retained in all summaries, even the shortest ones.
Main core ideas should be the key ideas that help the reader to understand the main topic. Depending on the type of documents, the definition of idea would be slightly different. For example, if the source document is a meeting note, the summary should include the main topic, the discussion, the result / final decision. The trivial details like greetings or small talks should not be included. If the source document is a novel, the summary should focus on main characters and important events rather than trivial description of the character or side events.
\item \textbf{Attribution}: the information in the summary can be accurately referred back to the  source documents. All the information in the summary should be an abstraction from the source documents. No additional information that can not be found in the source document should be included in the summary.
\item \textbf{Cohesion as a document}: Each summary should be an abstraction of the entire source document. All the information or ideas should be digested from different parts of the source document and combined into a new paragraph. Merely shortening a document paragraph by paragraph will not be considered as a good summary. Similarly, a bulletin-like document jumping from point to point also will not be considered as a good summary.

\end{itemize}

\paragraph{Annotation}

Our summarization structure is as follows:
The source text which is approximately 5,000 words long gets summarized three times:
The first summary is ~20\% of the original length of the doc. It should retain all the core ideas of the source document.
The second summary is ~10\% of the length of the source document. It should also retain all the core ideas of the source.
The third summary is short, it should be ~5\% of the source length. We understand there will be some information loss, but again, all the core ideas should be present in the summary.

There are two tasks related to evaluating the summary.

\paragraph{Task 1}

In this task, you need to rate the overall quality of the summaries regarding several aspects. Here is the detailed workflow:

\textbf{Step 1 Screening}

Please spend no more than 5 minutes skimming through the longest (20\%) summary and answer the question below.
\begin{itemize}
\item  Q1: Is it a cohesive text? Can you fully understand it?
\begin{itemize}
\item If NOT, reject the task completely.
\item If YES, continue with the following steps
\end{itemize}
\end{itemize}

\textbf{Step 2 Read the texts and take notes}

Please read the whole source document carefully and take notes in your own way. It could be highlighting the key points or jotting down the ideas in your own words or any means that can help you digest the document. While reading please do not skip any line. After reading and taking note, you should be able to identify several main core ideas or more (You can spot more main core ideas if the text is longer). Please continue to read the summary and identify if the main ideas also exist in the summaries. Now check how many ideas can be found in the summary.

(This procedure is to help rate the summaries more objectively, you are not required to submit the highlights or the notes.) 

\textbf{Step 3 Rate the summaries}

 Answer all the following questions with a 4-point scale:
\begin{itemize}
\item Q2a Check the attribution of the summary. Can all the information in the summary be attributed to the source text?

\begin{itemize} 
\item Give 4 points if yes, all the information can be directly attributed to the source text.
\item Give 3 points if mostly yes, only 1 idea seems to not be found in the source text.
\item Give 2 points if not really, more than 1 idea cannot be attributed to the source text.
\item Give 1 point if not, most ideas cannot be found in the source text and seem to be completely new.
\item Give 0 points if not, none of the ideas can be found in the source text.
\end{itemize}

\item  Q2b Check the coverage of main core ideas of the source text in the summary. Are all the main core ideas of the source document retained?
\begin{itemize} 
\item Give 4 points if yes, all the main core ideas of the source are retained.
\item Give 3 points if mostly yes, only 1 or 2 main core ideas are not found in the summary.
\item Give 2 points if not really, more than 2 main core ideas are not found in the summary.
\item Give 1 point if not, most main core ideas are not found in the summary.
\item Give 0 points if not, none of the ideas can be  found in the summary
\end{itemize}

\item  Q2c Check the conciseness of the summary. Is the summary short and clear without repetition and redundancy?
\begin{itemize} 
\item Give 4 points if yes, the summary is not wordy but clear.
\item Give 3 points if mostly yes, but 1 part is unnecessary. 
\item Give 2 points if not really, more than 1 part is unnecessary.
\item Give 1 point if not, the summary is lengthy and most passages can be omitted without losing the core ideas.
\item Give 0 points if not, the summary is lengthy and most passages can be omitted without losing the core ideas.
\end{itemize}

\item  Q2d Check the readability of the summary. Is the summary fluent and understandable?
\begin{itemize} 
\item Give 4 points if yes, the summary is fluent and understandable, and well written.
\item Give 3 points if mostly yes, but there is room for improvement.
\item Give 2 points if not really, not quite fluent and sometimes hard to understand.
\item Give 1 point if not, the summary is hard to read and understand.
\item Give 0 points if not, the summary is impossible to read and understand.
\end{itemize}

After evaluating all the aspects of the summary, please give an overall score of 0-10 on the quality of the summary.
\item Q3 Do you think it is a good summary? On a scale of 0-10, how would you rate the overall quality of the summary?
\begin{itemize} 
\item 10: The summary is perfect in every aspect
\item 8-9: The summary is considered good. It contains minor issues in certain aspects but it meets all requirements with room for improvement.
\item 6-7: The summary is moderate, it contains non-critical errors but to help the reader understand the source documents
\item 4-5: The summary is below acceptable level. It contains critical errors that could potentially mislead the reader. 
\item 2-3: The summary contains very limited information that is relevant to the source document
\item 0-1: The summary is barely readable and comprehensible or it barely contains relevant information to the source document.
\end{itemize}

Make sure you have answered all the questions for every summary.
\end{itemize}

\paragraph{Task 2}

You will be provided with 15 questions depending on the length of the documents. Please identify if the answer is directly stated, heavily implied, or logically entailed in the summary. You need to answer YES or NO only. 

\section{Summary Expansion Human Evaluation Guidelines}
\label{app:isum_guidelines}

\paragraph{Annotator proficiency requirements}
All annotators must meet ALL of the following requirements:
\begin{itemize} 
\item Native speaker of English AND
\item Language related degree holder or related professionals
\end{itemize}

\paragraph{Background Information}
What is a good summary expansion?
A good summary expansion should meet the following criteria:
\begin{itemize} 
\item \textbf{Coverage of main core ideas}: Main core ideas should be the key ideas covered in the original summary. During each expansion more details will be added, however the main core ideas should remain the same. For example, if the setting of the original summary is an office comedy, the ultimate text should not be an unrelated superhero movie screenplay.
\end{itemize}
What is a good long form text?
On top of coverage of main core ideas, a good summary expansion should also meet the following criteria:
Cohesion as a document: 

\begin{itemize} 
\item The ultimate document should be a stand-alone document so that the reader doesn’t need an additional document to understand the text. The ultimate document should be well-structured and formatted. For example, if the ultimate document is a screenplay, it should have clear scene sections, character direction and dialogs of characters. 
\item Non-repetitive and rich in details: Although the ultimate document is based on the summary, additional information / details is allowed. The ultimate document should not just repeat the core ideas.
\item Being Interesting: A good screenplay and novel should be able to capture the reader's attention and keep them engaged throughout the story / plot.  We are not searching for an Oscar winning novel / screenplay, as long as the plot makes sense and the added details serve the purpose of the story, it is considered as an interesting plot. For example, you can check the following questions depending on the story:
If it's a comedy, does it sound funny to you?
If it's a romance story, does it evoke the proper sentiment?
Are the added details aligned with the plot, or do they feel out of place?
… etc
\end{itemize}

\paragraph{Task 1 }

In this task, you need to rate the overall quality of the summary expansions regarding several aspects. Here is the detailed workflow:

\textbf{Step 1 Screening}

Please spend no more than 5 minutes skimming through the long form text and answer the question below.
\begin{itemize} 
\item Q1: Is it a cohesive text? Can you fully understand it?
\begin{itemize} 
\item If NOT, reject the task completely.
\item If YES, continue with the following steps
\end{itemize}
\end{itemize}

\textbf{Step 2 Read the texts and highlight key points}

Please read the original summary carefully and highlight the key points and make notes. Do not skip any line. Continue to read other summaries and long form text, highlighting the key points that are the same as the original summary

(This procedure is to help rating the summaries more objectively, you are not required to submit the highlights or the notes.)

\textbf{Step 3 Rate the long form text}

Answer all the following questions with a 4-point scale separately for the long form text:

\begin{itemize} 
\item Q2a Check the coverage of main core ideas. Are all the core concepts of the original summary retained?

\begin{itemize} 
\item Give 4 points if yes, all the main core ideas are retained.
\item Give 3 points if mostly yes, only 1 or 2 core ideas are lost.
\item Give 2 points if not really, more than 2 ideas are lost.
\item Give 1 point if not, most ideas are lost.
\item Give 0 points if not, none of the ideas can be found in the source text.
\end{itemize} 

\item Q2b Check the cohesion of text. Is it well structured? Does it contain all the necessary components? (Scene description, dialog, main characters, etc) Does it flow logically and maintain consistency?

\begin{itemize} 
\item Give 4 points if yes, it is a well structure screenplay / novel 
\item Give 3 points if mostly yes, but 1 part is missing.
\item Give 2 points if not really, more than 1 part is missing.
\item Give 1 point if text does not follow the structure of a screenplay / novel
\item Give 0 points if not, the text doesn't not read as a cohesive text at all
\end{itemize} 

\item Q2c Check the richness in details. Does it contain enough details?

\begin{itemize} 
\item Give 4 points if yes, the text contains a lot of details.
\item Give 3 points if yes, the text contains details but has room for improvement.
\item Give 2 points if not really, the text contains limited details.
\item Give 1 point if not, the text contains very few details
\item Give 0 points if not, the text does not provide any additional details at all.
\end{itemize} 

\item Q2d Check the creativity. Does the added details novel and original while being relevant to the core main ideas?

\begin{itemize} 
\item Give 4 points if yes, all of the added details are novel and original
\item Give 3 points if yes, most of the added details are novel and original.
\item Give 2 points if not really, only some of the added details are novel and original.
\item Give 1 point if no, very few added details are repetitive.
\item Give 0 points if no, no added details are novel and original .
\end{itemize} 

\item Q2e Check the non-repetitiveness. Does it repeat a lot?

\begin{itemize} 
\item Give 4 points if no, all of the details are unique and different from each other
\item Give 3 points if no, most of the details are unique, only one is repeated
\item Give 2 points if not really, some of the details are repetitive
\item Give 1 point if yes, most the added details are repetitive
\item Give 0 points if not, all the added details are repetitive
\end{itemize} 

\item Q2f Rate the story plot. How interesting is it to you? Is it engaging and compelling?

\begin{itemize} 
\item Give 4 points if the text is very interesting.
\item Give 3 points if the text is quite interesting.
\item Give 2 points if the text is somewhat interesting.
\item Give 1 point if the text is only slightly interesting.
\item Give 0 points if the text is dull and not interesting at all.
\end{itemize} 

After evaluating all the aspects of the expanded text, please give an overall score of 0-10 on the quality of the expanded text.
\item Q3 Do you think expanded text is well written? Do you think it is a good read?  On a scale of 0-
10, how would you rate the overall quality?
\begin{itemize} 
\item 10: The text is perfect in every aspects
\item 8-9: The text is considered good. It contains minor issues in certain aspects but it meets all requirements with room for improvement.
\item 6-7: The text is moderate, it contains non-critical errors but to help the reader understand the source documents
\item 4-5: The text is below acceptable level. It contains critical errors that cause trouble to read
\item 2-3: The text contains very limited information that is relevant to the source summary
\item 0-1: The text is barely readable and comprehensible or it barely contains relevant information to the source summary.
\end{itemize} 
\end{itemize} 
Make sure you have answered all the questions for every expanded summary and long form text.

\paragraph{Task 2}

You will be provided with around 15 questions depending on the length of the documents. 
You need to answer YES or NO to each QA set. Answer YES only when the answer is directly stated, heavily implied, or logically entailed in the text.

\section{Prompting Details}
\label{app:prompting}

The prompts contain three parts: General guideline, domain-specific prompts, and input context. The general guideline adapts the human guidelines (Appendix~\ref{human:summguide}) for the summarization and summary expansion, while the domain specific prompts give extra information about the domain as instructions of expected output. In the prompt template below, the general guideline is provided, \texttt{\{\{domain-X\}\}} denotes the domain-specific prompt. \texttt{\{\{input\}\}} is for the input document for the summarization task and human summaries for the summary expansion task.

\paragraph{Prompt for summarization}

\begin{small}
\begin{verbatim}
"""
You are a professional editor and reader. 
You are reading a {{domain}} 
{{domain-meta}}. 
The {{domain}} starts with [START] 
and ends with [END]. 
After you have finished reading, please 
provide a summary of the {{domain}}. 
{{domain-expect}}.
Make sure the summary has 
{{len(input) * ratio + 200}} words or 
less. 
[START]
{{input}}
[END].
Write at least {{len(input) * ratio}}
words.
"""
\end{verbatim}
\end{small}

\paragraph{Prompt for summary expansion}
\begin{small}
\begin{verbatim}
"""
You are a professional editor and reader. 
You are reading a summary of {{domain}} 
{{domain-meta}}. 
The {{domain}} starts with [START] 
and ends with [END]. 
After you have finished reading, write
a well-structured, consistent {{domain}}
that extends the summary. 
{{domain-expect-expand}}.
[START]
{{input}}
[END].
Write at least {{len(source)}} words.
"""
\end{verbatim}
\end{small}

The model-specific prompts for each domain are listed below. Note that not all domains have the prompt template for summary expansion.

\begin{itemize}

    \item \textbf{BookSum:}
\begin{small}
\begin{verbatim}{{domain}}: "book chapter"
{{domain-meta}}: """about the book
  [BOOK-TITLE], chapter [CHAP-NO], 
  title [[CHAP-TITLE]]."""
{{domain-expect}}: ""
{{domain-expect-expand}}: """Please
  keep the main plot and characters
  if found in the summary.
""".
\end{verbatim}
\end{small}

    \item \textbf{LexGLUE:}
\begin{small}
\begin{verbatim}{{domain}}: "legal document"
{{domain-meta}}: ""
{{domain-expect}}: """Keep the main
  ideas and terms in the document.
"""
\end{verbatim}
\end{small}

    \item \textbf{SQuALITY:}
\begin{small}
\begin{verbatim}{{domain}}: "short story"
{{domain-meta}}: ""
{{domain-expect}} : """Keep the main
  character names and narratives
  of the story.
"""
{{domain-expect-expand}}: (same)
\end{verbatim}
\end{small}

    \item \textbf{Seahorse:}
\begin{small}
\begin{verbatim}{{domain}}: "news article"
{{domain-meta}}: ""
{{domain-expect}}: ""
\end{verbatim}
\end{small}

    \item \textbf{FacetSum:}
\begin{small}
\begin{verbatim}{{domain}}: "academic article"
{{domain-meta}}: "about [TITLE]"
{{domain-expect}}: """Keep the
  structure of sections [SECTIONS]
"""
{{domain-expect-expand}}: (same)
\end{verbatim}
\end{small}

    \item \textbf{JRC-Acquis:}
\begin{small}
\begin{verbatim}{{domain}}: "document"
{{domain-meta}}: "from  European Commision"
{{domain-expect}}: ""
\end{verbatim}
\end{small}

    \item \textbf{MultiUN:}
\begin{small}
\begin{verbatim}{{domain}}: "document"
{{domain-meta}}: "from United Nation"
{{domain-expect}}: ""
\end{verbatim}
\end{small}

    \item \textbf{GovReport:}
\begin{small}
\begin{verbatim}{{domain}}: "government report"
{{domain-meta}}: ""
{{domain-expect}}: ""
\end{verbatim}
\end{small}

    \item \textbf{Wikipedia:}
\begin{small}
\begin{verbatim}{{domain}}: "Wikipedia article"
{{domain-meta}}: "about [TITLE]"
{{domain-expect}}: ""
\end{verbatim}
\end{small}

    \item \textbf{Summscreen:}
\begin{small}
\begin{verbatim}{{domain}}: "screenplay"
{{domain-meta}}: "about [TITLE]"
{{domain-expect}}: """Keep the main
  plot and characters in the screenplay
"""
{{domain-expect-expand}}: """Keep the 
  main plot and characters in the
  summary.
  Write in the dialogue form with 
  multiple utterances 
"""
\end{verbatim}
\end{small}


\section{Detailed Results}
\label{app:detailedresults}
Complementing Table \ref{tabl:isummarization} from Section \ref{sec:experiments}, Tables~\ref{tabl:isum_all_10} and \ref{tabl:isum_all_5} show results of different models in the summary expansion tasks on a different level of summary and expansion. Tables \ref{tab:isumm5}, \ref{tab:isumm10} and \ref{tab:isumm20} show the detailed results breaked down by domain on the summary expansion task given the summaries (5\%, 10\% and 20\%) and expanding by a respective factor of (20, 10, 5). It is shown that, despite having repeated instructions on the length, all models behave greatly differently in different domains. In particular, GPT-4o-mini can generate longer scientific/technical texts, but struggle to generate longer texts in conversational texts without sacrifying the qualities. On the other hand, medium-sized models such as \llama 3.1-8B generate more texts consistently across domains, but at a higher repetition. 

\begin{table*}[ht!]
\centering
\scriptsize
\sisetup{table-format = 3.2}
\begin{tabular}{@{}p{2.2cm}cccccccc@{}}
\toprule
Output & \%WC & \textsc{REP-3(\(\downarrow\))} & \textsc{CoLA\(\uparrow\)} & \textsc{COH-2\(\uparrow\)} & \textsc{AVG\(\uparrow\)}  & \textsc{HE\(\uparrow\)} \\
\midrule

GPT-4o-mini & \textbf{4.789} & 0.623 & \textbf{0.910} & 0.564 & 0.450 & 67.617 \\
\addlinespace[0.3em]

\llama 3.1-70B & 2.807 & 1.830 & 0.913 & 0.625 & 0.391 & 54.344  \\
\addlinespace[0.3em]

\llama 3.1-8B & 1.788 & 0.781 & 0.906 & 0.752 & 0.501 & 42.572 \\

\addlinespace[0.5em]

\bottomrule
\end{tabular}%
\caption{Performance on the summary expansion task by a factor of 10, given the 10\% summary input.}
\label{tabl:isum_all_10}
\end{table*}

\begin{table*}[ht!]
\centering
\scriptsize
\sisetup{table-format = 3.2}
\begin{tabular}{@{}p{2.2cm}cccccccc@{}}
\toprule
Output & \%WC & \textsc{REP-3(\(\downarrow\))} & \textsc{CoLA\(\uparrow\)} & \textsc{COH-2\(\uparrow\)} & \textsc{AVG\(\uparrow\)}  & \textsc{HE\(\uparrow\)} \\
\midrule

GPT-4o-mini & \textbf{7.960} & 0.592 & 0.909 & 0.574 & 0.455 & 65.951 \\
\addlinespace[0.3em]

\llama 3.1-70B & 2.296 & 0.631 & 0.871 & 0.720 & 0.488 & 51.539  \\
\addlinespace[0.3em]

\llama 3.1-8B & 3.805 & 0.720 & 0.967 & 0.862 & 0.562 & 45.141 \\

\addlinespace[0.5em]

\bottomrule
\end{tabular}%
\caption{Performance on the summary expansion task by a factor of 20, given the 5\% summary input.}
\label{tabl:isum_all_5}
\end{table*}

\begin{table*}[htb]
\centering
\fontsize{10}{10}\selectfont
\resizebox{1.0\textwidth}{!}{%
\begin{tabular}{@{}p{2.1cm}rcccccccc@{}}
\toprule
\textsc{Dataset} & Model & \multicolumn{1}{l}{\textsc{\% WC}} & \textsc{REP-3(\(\downarrow\))} & \textsc{CoLA\(\uparrow\)} & \textsc{COH-2\(\uparrow\)} & \textsc{AVG\(\uparrow\)} & \textsc{HE\(\uparrow\)} \\
\midrule
\addlinespace[0.5em]
\multirow{3}{8pt}{\textbf{BookSum}} & GPT-4o-mini & 7.116 & 0.420 & 0.953 & 0.701 & 0.523  & 82.421  \\
& \llama 3.1-70B & 3.307 & 0.631 & 0.931 & 0.863 & 0.556  & 66.252  \\
& \llama 3.1-8B & 5.070 & 2.170 & 0.975 & 0.787 & 0.442  & 60.393 \\
\addlinespace[0.5em]

\multirow{3}{8pt}{\textbf{SQuALITY}} & GPT-4o-mini & 7.335 & 0.398 & 0.954 & 0.695 & 0.523  & 55.620 \\
& \llama 3.1-70B & 3.010 & 0.530 & 0.946 & 0.769 & 0.536  & 47.703  \\
& \llama 3.1-8B & 3.166 & 0.624 & 0.947 & 0.795 & 0.539 & 42.532 \\

\addlinespace[0.5em]

\multirow{3}{8pt}{\textbf{FacetSum}} & GPT-4o-mini & 9.348 & 0.635 & 0.936 & 0.593 & 0.467  & 59.238 \\
& \llama 3.1-70B & 0.253 & 0.792 & 0.769 & 0.816 & 0.475 & 47.145 \\
& \llama 3.1-8B & 0.369 & 0.998 & 0.843 & 0.841 & 0.495 & 27.660 \\

\addlinespace[0.5em]

\multirow{3}{8pt}{\textbf{Summscreen}} & GPT-4o-mini & 8.042 & 0.914 & 0.794 & 0.309 & 0.307  & 74.779 \\ 
& \llama 3.1-70B & 2.615 & 0.572 & 0.836 & 0.432 & 0.385 & 50.914 \\
& \llama 3.1-8B & 3.370 & 0.712 & 0.840 & 0.474 & 0.391 & 44.121 \\

\addlinespace[0.5em]
\bottomrule
\end{tabular}%
}
\caption{Performance on the summary expansion task by a factor of 20, given the 5\% summary input, per dataset.}
\label{tab:isumm5}
\end{table*}

\begin{table*}[htb]
\centering
\fontsize{10}{10}\selectfont
\resizebox{1.0\textwidth}{!}{%
\begin{tabular}{@{}p{2.1cm}rcccccccc@{}}
\toprule
\textsc{Dataset} & Model & \multicolumn{1}{l}{\textsc{\% WC}} & \textsc{REP-3(\(\downarrow\))} & \textsc{CoLA\(\uparrow\)} & \textsc{COH-2\(\uparrow\)} & \textsc{AVG\(\uparrow\)} & \textsc{HE\(\uparrow\)} \\
\midrule
\addlinespace[0.5em]
\multirow{3}{8pt}{\textbf{BookSum}} & GPT-4o-mini & 5.155 & 0.443 & 0.955 & 0.689 & 0.518  & 82.421 \\
& \llama 3.1-70B & 3.596 & 1.453 & 0.981 & 0.817 & 0.503 & 74.280  \\
& \llama 3.1-8B & 2.451 & 0.778 & 0.950 & 0.859 & 0.551 & 60.163 \\

\addlinespace[0.5em]

\multirow{3}{8pt}{\textbf{SQuALITY}} & GPT-4o-mini & 3.531 & 0.418 & 0.952 & 0.665 & 0.511 & 56.867 \\
& \llama 3.1-70B & 3.104 & 1.098 & 0.957 & 0.718 & 0.485 & 50.855  \\
& \llama 3.1-8B & 2.056 & 0.659 & 0.951 & 0.813 & 0.544 & 38.008  \\

\addlinespace[0.5em]

\multirow{3}{8pt}{\textbf{FacetSum}} & GPT-4o-mini & 5.220 & 0.675 & 0.939 & 0.573 & 0.459 & 62.472 \\
& \llama 3.1-70B & 0.504 & 3.005 & 0.949 & 0.707 & 0.352 & 33.248 \\
& \llama 3.1-8B & 0.356 & 0.952 & 0.882 & 0.860 & 0.517 & 29.713 \\

\addlinespace[0.5em]

\multirow{3}{8pt}{\textbf{Summscreen}} & GPT-4o-mini & 5.251 & 0.954 & 0.794 & 0.331 & 0.311 & 68.707 \\ 
& \llama 3.1-70B & 4.023 & 1.762 & 0.766 & 0.259 & 0.224 & 58.992 \\
& \llama 3.1-8B & 2.290 & 0.736 & 0.843 & 0.475 & 0.390 & 42.404 \\

\addlinespace[0.5em]
\bottomrule
\end{tabular}%
}
\caption{Performance on the 10\% summary expansion task per dataset.}
\label{tab:isumm10}
\end{table*}

\begin{table*}[htb]
\centering
\fontsize{10}{10}\selectfont
\resizebox{1.0\textwidth}{!}{%
\begin{tabular}{@{}p{2.1cm}rcccccccc@{}}
\toprule
\textsc{Dataset} & Model & \multicolumn{1}{l}{\textsc{\% WC}} & \textsc{REP-3(\(\downarrow\))} & \textsc{CoLA\(\uparrow\)} & \textsc{COH-2\(\uparrow\)} & \textsc{AVG\(\uparrow\)} & \textsc{HE\(\uparrow\)} 
& \textsc{Hum\(\uparrow\)} \\
\midrule
\addlinespace[0.5em]
\multirow{3}{8pt}{\textbf{BookSum}} & GPT-4o-mini & 0.641 & 0.459 & 0.960 & 0.739 & 0.536  & 87.161  &  6.691 \\
& \llama 3.1-70B & 1.539 & 0.650 & 0.867 & 0.842 & 0.526  & 50.969 & 5.123  \\
& \llama 3.1-8B & 1.695 & 0.736 & 0.936 & 0.861 & 0.550  & 53.692 & 5.160 \\
\addlinespace[0.5em]

\multirow{3}{8pt}{\textbf{SQuALITY}} & GPT-4o-mini & 3.014 & 0.513 & 0.961 & 0.738 & 0.532  & 60.385 & 6.320  \\
& \llama 3.1-70B & 1.107 & 0.582 & 0.952 & 0.780 & 0.539  & 37.775 &  5.434  \\
& \llama 3.1-8B & 1.351 & 0.735 & 0.955 & 0.817 & 0.542 & 35.017 & 5.360  \\

\addlinespace[0.5em]

\multirow{3}{8pt}{\textbf{FacetSum}} & GPT-4o-mini & 0.502 & 0.642 & 0.954 & 0.640 & 0.488  & 70.061 &  7.693  \\
& \llama 3.1-70B & 0.138 & 0.705 & 0.874 & 0.877 & 0.537  & 26.587 &3.453 \\
& \llama 3.1-8B & 0.205 & 0.935 & 0.871 & 0.871 & 0.518 & 28.462 & 4.173 \\

\addlinespace[0.5em]

\multirow{3}{8pt}{\textbf{Summscreen}} & GPT-4o-mini & 3.569 & 1.215 & 0.776 & 0.319 & 0.284  & 65.977 &  5.000  \\ 
& \llama 3.1-70B & 1.449 & 0.782 & 0.814 & 0.500 & 0.386  & 41.466 &3.800 \\
& \llama 3.1-8B & 1.495 & 0.828 & 0.851 & 0.568 & 0.418 & 36.493 &4.480\\

\addlinespace[0.5em]
\bottomrule
\end{tabular}%
}
\caption{Performance on the summary expansion task by a factor of 5, given the 20\% summary input.}
\label{tab:isumm20}
\end{table*}

The results of the summarization task of different levels are detailed in tables~\ref{tab:summarization-5},~\ref{tab:summarization-10} and~\ref{tab:summarization-20}. According to human evaluation, the best performing result in Table \ref{tab:summarization-5} is with GPT-4o-mini in the wikipedia domain and in Table \ref{tab:summarization-10} is with the same model in the legal domain (LexGlue). For table \ref{tab:summarization-20}, the best results are with human output in the conversational domain (Summscreen).

\begin{table*}[tb]
\centering
\fontsize{10}{10}\selectfont
\sisetup{table-format = 3.2}
\resizebox{\textwidth}{!}{%
\begin{tabular}{@{}p{2.1cm}rccccccccc@{}}
\toprule
\textsc{Dataset} & Model & \multicolumn{1}{l}{\textsc{R-L(\(\uparrow\))}} & \textsc{REP-3(\(\downarrow\))} & \textsc{CoLA\(\uparrow\)} & \textsc{COH-2\(\uparrow\)} & \textsc{SH-4\(\uparrow\)} & \textsc{SH-5\(\uparrow\)} & \textsc{AVG\(\uparrow\)} &  \textsc{HE\(\uparrow\)} 
&  \textsc{Hum\(\uparrow\)}  \\
\midrule
\addlinespace[0.3em]
\rowcol\multicolumn{11}{c}{\textbf{LCFO.5\%}} \\
\addlinespace[0.5em]
\multirow{4}{8pt}{\textbf{LexGLUE}} 
& Human& n/a & 0.258	& 0.930	& 0.807	& 0.617	& 0.339	& 0.528 & 46.690 & 6.360 \\
& GPT-4o-mini & 0.342 & 0.407	& 0.956	& 0.688	& 0.657	& 0.500	& 0.479  & 78.916 &  7.747\\
& \llama 3.1-70B & 0.386 & 0.415 & 0.954 & 0.875 & 0.625 & 0.369 & 0.482 & 63.280 &  6.987\\
& \llama 3.1-8B & 0.378 & 0.471 & 0.972 & 0.879 & 0.617 & 0.383 & 0.476  & 59.455 &  6.907\\

\addlinespace[0.5em]

\multirow{3}{8pt}{\textbf{BookSum}} 
& Human& n/a & 	0.226	& 0.913	& 0.762	& 0.572 & 	0.315 & 	0.503&  71.006 &  6.691\\
& GPT-4o-mini & 0.302 & 0.257 &	0.977 &	0.857 &	0.599 &	0.485 &	0.532  & 93.168 &  6.815 \\
& \llama 3.1-70B & 0.377 & 0.362 & 0.976 & 0.846 & 0.578 & 0.374 & 0.483 & 76.871 &  6.272 \\
& \llama 3.1-8B & 0.372 & 0.400 & 0.973 & 0.846 & 0.581 & 0.347 & 0.469  & 72.999 &  6.049 \\
\addlinespace[0.5em]

\multirow{3}{8pt}{\textbf{SQuALITY}} 
& Human& n/a & 	0.263	& 0.922	& 0.760 & 	0.520 &	0.334	& 0.497  &  33.534 &  5.173\\
& GPT-4o-mini & 0.285 & 0.284 & 0.980 & 0.841 & 0.548 & 0.375 & 0.492  & 74.618  &  6.600 \\
& \llama 3.1-70B & 0.340 & 0.472 & 0.961 & 0.802 & 0.463 & 0.201 & 0.391 & 64.237  &  5.227  \\
& \llama 3.1-8B & 0.339 & 0.535 & 0.968 & 0.819 & 0.488 & 0.233 & 0.395 & 57.288  &  5.827 \\

\addlinespace[0.5em]

\multirow{3}{8pt}{\textbf{FacetSum}} 
& Human& n/a & 	0.260 & 	0.945 & 	0.835 & 	0.691 & 	0.436 & 	0.571 & 57.456 & 7.053 \\
& GPT-4o-mini & 0.404 & 0.354 & 0.921 & 0.568 & 0.682 & 0.524 & 0.468 & 73.968 &  7.434  \\
& \llama 3.1-70B & 0.412 & 0.387 & 0.962 & 0.884 & 0.696 & 0.508 & 0.533& 67.585 &  6.213 \\
& \llama 3.1-8B & 0.419 & 0.425 & 0.967 & 0.888 & 0.704 & 0.518 & 0.530  & 69.176 &  6.733 \\

\addlinespace[0.5em]

\multirow{3}{8pt}{\textbf{JRC-Acquis}} 
& Human& n/a & 0.247 & 	0.949 & 	0.849 & 	0.672 & 	0.464 & 	0.577 & 52.092  & 7.180 \\
& GPT-4o-mini & 0.352 & 0.383 & 0.952 & 0.539 & 0.682 & 0.593 & 0.477  & 82.239  &  7.347 \\
& \llama 3.1-70B & 0.390 & 0.424 & 0.942 & 0.883 & 0.690 & 0.470 & 0.512  & 60.948 &  6.306 \\
& \llama 3.1-8B & 0.368 & 0.427 & 0.945 & 0.882 & 0.673 & 0.449 & 0.504 & 59.209 &  6.514\\

\addlinespace[0.5em]

\multirow{3}{8pt}{\textbf{MultiUN}} 
& Human& n/a & 	0.255 & 	0.927 & 	0.862 & 	0.592 & 	0.276 & 	0.521 & 44.466 & 6.861 \\
& GPT-4o-mini & 0.352 & 0.364 & 0.968 & 0.549 & 0.630 & 0.528 & 0.462 & 76.639 &  7.347 \\
& \llama 3.1-70B & 0.402 & 0.400 & 0.955 & 0.903 & 0.618 & 0.303 & 0.476 & 76.121 &  6.611 \\
& \llama 3.1-8B & 0.378 & 0.443 & 0.965 & 0.907 & 0.608 & 0.320 & 0.471 & 59.683 &  6.806\\

\addlinespace[0.5em]

\multirow{3}{8pt}{\textbf{Wikipedia}} 
& Human& n/a & 0.246 & 	0.961 & 	0.810 & 	0.664 & 	0.246 & 	0.527 &  68.484 & 6.893 \\
& GPT-4o-mini & 0.341 & 0.332 & 0.974 & 0.756 & 0.693 & 0.423 & 0.503 & 80.633  &  7.754 \\
& \llama 3.1-70B & 0.382 & 0.405 & 0.968 & 0.821 & 0.660 & 0.299 & 0.469 & 59.334 &  6.551 \\
& \llama 3.1-8B & 0.379 & 0.439 & 0.963 & 0.839 & 0.672 & 0.282 & 0.463 & 59.259 &  5.841\\

\addlinespace[0.5em]

\multirow{3}{8pt}{\textbf{GovReport}} 
& Human& n/a & 	0.226 & 	0.958 & 	0.803 & 	0.639 & 	0.336 & 	0.538 & 35.157 &  6.720\\
& GPT-4o-mini & 0.340 & 0.333 & 0.978 & 0.722 & 0.696 & 0.538 & 0.520 & 81.420  & 7.280 \\
& \llama 3.1-70B & 0.407 & 0.363 & 0.973 & 0.870 & 0.651 & 0.430 & 0.512 & 54.626 &  6.080 \\
& \llama 3.1-8B & 0.407 & 0.353 & 0.971 & 0.855 & 0.620 & 0.354 & 0.489 & 52.231 & 6.44 \\

\addlinespace[0.5em]

\multirow{3}{8pt}{\textbf{Summscreen}} 
& Human& n/a & 0.243 & 	0.927 & 	0.739 & 	0.532 & 	0.384 & 	0.507 & 62.003 & 7.040 \\
& GPT-4o-mini & 0.294 & 0.289 & 0.984 & 0.832 & 0.514 & 0.346 & 0.478 & 60.347 &  6.627  \\
& \llama 3.1-70B & 0.390 & 0.328 & 0.985 & 0.849 & 0.523 & 0.259 & 0.458 & 70.638  &  6.173 \\
& \llama 3.1-8B & 0.375 & 0.373 & 0.976 & 0.854 & 0.526 & 0.266 & 0.450 & 69.116 &  5.667\\

\addlinespace[0.5em]

\multirow{3}{8pt}{\textbf{Seahorse}} 
& Human& n/a & 	0.213 & 	0.950 & 	0.819 & 	0.651 & 	0.440 & 	0.563 & 51.057 &  6.200\\
& GPT-4o-mini & 0.295 & 0.279 & 0.985 & 0.832 & 0.647 & 0.556 & 0.548 & 67.220 &  7.613 \\
& \llama 3.1-70B & 0.352 & 0.382 & 0.965 & 0.842 & 0.661 & 0.434 & 0.504 & 65.295 &  6.293 \\
& \llama 3.1-8B & 0.354 & 0.369 & 0.978 & 0.846 & 0.649 & 0.472 & 0.515 & 65.893 &  6.427\\

\addlinespace[0.5em]
\bottomrule
\end{tabular}%
}
\caption{Performance on the 5\% summarization task per dataset.}
\label{tab:summarization-5}
\end{table*}


\begin{table*}[htb]
\centering
\fontsize{10}{10}\selectfont
\sisetup{table-format = 3.2}
\resizebox{\textwidth}{!}{%
\begin{tabular}{@{}p{2.1cm}rccccccccc@{}}
\toprule
\textsc{Dataset} & Model & \multicolumn{1}{l}{\textsc{R-L(\(\uparrow\))}} & \textsc{REP-3(\(\downarrow\))} & \textsc{CoLA\(\uparrow\)} & \textsc{COH-2\(\uparrow\)} & \textsc{SH-4\(\uparrow\)} & \textsc{SH-5\(\uparrow\)} & \textsc{AVG\(\uparrow\)}  & \textsc{HE\(\uparrow\)} 
& \textsc{Hum\(\uparrow\)} \\
\midrule
\addlinespace[0.3em]
\rowcol\multicolumn{11}{c}{\textbf{LCFO.10\%}} \\
\addlinespace[0.5em]
\addlinespace[0.5em]
\multirow{4}{8pt}{\textbf{LexGLUE}} & Human& n/a & 0.351 & 	0.940 & 	0.829 & 	0.660 & 	0.362 & 	0.544 &  62.141& 7.387 \\
& GPT-4o-mini & 0.419 & 0.494 & 0.947 & 0.599 & 0.633 & 0.500 & 0.437 & 80.094 &  8.120 \\
& \llama 3.1-70B & 0.452 & 0.566 & 0.942 & 0.882 & 0.625 & 0.397 & 0.456 & 59.666 &  7.080  \\
& \llama 3.1-8B & 0.439 & 0.641 & 0.967 & 0.876 & 0.621 & 0.376 & 0.440 & 59.349 &  7.200 \\

\addlinespace[0.5em]

\multirow{3}{8pt}{\textbf{BookSum}} 
& Human& n/a & 0.278 & 	0.907 & 	0.757 & 	0.610 & 	0.342 & 	0.512 & 83.776 &  7.601\\
& GPT-4o-mini & 0.327 & 0.308 & 0.978 & 0.858 & 0.578 & 0.442 & 0.510 & 94.867  &  7.062 \\
& \llama 3.1-70B & 0.427 & 0.456 & 0.967 & 0.835 & 0.573 & 0.335 & 0.451 & 82.114 &  6.469  \\
& \llama 3.1-8B & 0.415 & 0.511 & 0.966 & 0.844 & 0.551 & 0.313 & 0.432 & 73.681  &  6.420 \\

\addlinespace[0.5em]

\multirow{3}{8pt}{\textbf{SQuALITY}} 
& Human& n/a & 	0.313 & 	0.917 & 	0.773 & 	0.548 & 	0.312 & 	0.497 & 51.501 &  6.000\\
& GPT-4o-mini & 0.327 & 0.329 & 0.975 & 0.819 & 0.525 & 0.341 & 0.466 & 76.441  & 6.467  \\
& \llama 3.1-70B & 0.382 & 0.367 & 0.974 & 0.836 & 0.518 & 0.320 & 0.456 & 46.731 & 4.613  \\
& \llama 3.1-8B & 0.373 & 0.406 & 0.979 & 0.856 & 0.526 & 0.324 & 0.456 & 50.129 & 5.280 \\

\addlinespace[0.5em]

\multirow{3}{8pt}{\textbf{FacetSum}} 
& Human& n/a &  0.328 & 	0.942 & 	0.840 & 	0.710 & 	0.425 & 	0.570 & 69.570& 7.680 \\
& GPT-4o-mini & 0.461 & 0.409 & 0.945 & 0.658 & 0.666 & 0.506 & 0.473 & 78.381  & 7.882  \\
& \llama 3.1-70B & 0.455 & 0.538 & 0.934 & 0.890 & 0.696 & 0.527 & 0.502 & 63.663 & 6.501 \\
& \llama 3.1-8B & 0.449 & 0.547 & 0.954 & 0.892 & 0.698 & 0.496 & 0.499 & 63.768 & 6.987 \\

\addlinespace[0.5em]

\multirow{3}{8pt}{\textbf{JRC-Acquis}} 
& Human& n/a & 0.339 & 	0.959 & 	0.845 & 	0.702 & 	0.512 & 	0.590 &  61.618& 7.680 \\
& GPT-4o-mini & 0.433 & 0.479 & 0.947 & 0.548 & 0.668 & 0.543 & 0.445 & 81.398 & 7.819  \\
& \llama 3.1-70B & 0.440 & 0.579 & 0.922 & 0.888 & 0.662 & 0.455 & 0.470 & 52.570& 6.542  \\
& \llama 3.1-8B & 0.443 & 0.586 & 0.938 & 0.859 & 0.673 & 0.441 & 0.465 & 49.986 & 7.02 \\

\addlinespace[0.5em]

\multirow{3}{8pt}{\textbf{MultiUN}} 
& Human& n/a & 0.312 & 	0.942 & 	0.871 & 	0.612 & 	0.334 & 	0.539 & 57.357  & 7.902 \\
& GPT-4o-mini & 0.422 & 0.455 & 0.961 & 0.629 & 0.621 & 0.518 & 0.455 & 74.459  & 7.875 \\
& \llama 3.1-70B & 0.447 & 0.546 & 0.912 & 0.914 & 0.622 & 0.329 & 0.446 & 52.920 & 6.903 \\
& \llama 3.1-8B & 0.446 & 0.557 & 0.950 & 0.900 & 0.606 & 0.295 & 0.439 & 55.186& 7.014 \\

\addlinespace[0.5em]

\multirow{3}{8pt}{\textbf{Wikipedia}} 
& Human& n/a & 	0.286 & 	0.969 & 	0.812 & 	0.723 & 	0.286 & 	0.547 & 78.316 &  7.747\\
& GPT-4o-mini & 0.388 & 0.428 & 0.963 & 0.640 & 0.690 & 0.446 & 0.462 & 78.149 & 7.696   \\
& \llama 3.1-70B & 0.445 & 0.557 & 0.931 & 0.830 & 0.670 & 0.278 & 0.430 & 57.310 & 6.681  \\
& \llama 3.1-8B & 0.444 & 0.489 & 0.963 & 0.822 & 0.691 & 0.322 & 0.462 & 58.038 & 6.246 \\

\addlinespace[0.5em]

\multirow{3}{8pt}{\textbf{GovReport}} 
& Human& n/a & 0.296 & 	0.956 & 	0.815 & 	0.670 & 	0.361 & 	0.548 & 52.627 &  6.720\\
& GPT-4o-mini & 0.402 & 0.420 & 0.972 & 0.639 & 0.683 & 0.529 & 0.480 & 81.374 & 7.72  \\
& \llama 3.1-70B & 0.454 & 0.511 & 0.923 & 0.874 & 0.667 & 0.406 & 0.472 & 49.294 & 6.57  \\
& \llama 3.1-8B & 0.459 & 0.498 & 0.972 & 0.871 & 0.650 & 0.415 & 0.482 & 52.712 & 6.987 \\

\addlinespace[0.5em]

\multirow{3}{8pt}{\textbf{Summscreen}} 
& Human& n/a & 	0.308 & 	0.930 & 	0.732 & 	0.557 & 	0.414 & 	0.514 & 68.971 &  7.733\\
& GPT-4o-mini & 0.314 & 0.364 & 0.974 & 0.778 & 0.511 & 0.345 & 0.449 & 61.149 & 6.667  \\
& \llama 3.1-70B & 0.417 & 0.436 & 0.984 & 0.849 & 0.505 & 0.281 & 0.437 & 49.294& 6.413  \\
& \llama 3.1-8B & 0.402 & 0.501 & 0.987 & 0.855 & 0.506 & 0.295 & 0.428 & 62.788 & 6.067  \\

\addlinespace[0.5em]

\multirow{3}{8pt}{\textbf{Seahorse}} 
& Human& n/a & 	0.271 & 	0.952 & 	0.811 & 	0.651 & 	0.518 & 	0.576 & 60.999 & 7.067 \\
& GPT-4o-mini & 0.353 & 0.353 & 0.977 & 0.786 & 0.632 & 0.542 & 0.517 & 72.321  & 7.720 \\
& \llama 3.1-70B & 0.417 & 0.491 & 0.963 & 0.831 & 0.660 & 0.480 & 0.489 & 60.241 & 6.440  \\
& \llama 3.1-8B & 0.402 & 0.474 & 0.968 & 0.839 & 0.642 & 0.476 & 0.490 & 60.415 & 7.080 \\

\addlinespace[0.5em]
\bottomrule
\end{tabular}%
}
\caption{Performance on the 10\% summarization task per dataset}
\label{tab:summarization-10}
\end{table*}


\begin{table*}[htb]
\centering
\fontsize{10}{10}\selectfont
\sisetup{table-format = 3.2}
\resizebox{\textwidth}{!}{%
\begin{tabular}{@{}p{2.5cm}rccccccccc@{}}
\toprule
\textsc{Dataset} & Model & \multicolumn{1}{l}{\textsc{R-L(\(\uparrow\))}} & \textsc{REP-3(\(\downarrow\))} & \textsc{CoLA\(\uparrow\)} & \textsc{COH-2\(\uparrow\)} & \textsc{SH-4\(\uparrow\)} & \textsc{SH-5\(\uparrow\)} & \textsc{AVG\(\uparrow\)}  & \textsc{HE\(\uparrow\)} 
&\textsc{Hum\(\uparrow\)}  \\
\midrule
\addlinespace[0.3em]
\rowcol\multicolumn{11}{c}{\textbf{LCFO.20\%}} \\
\addlinespace[0.3em]
\multirow{4}{8pt}{\textbf{LexGLUE}}  
& Human& n/a & 0.455 & 	0.940 & 	0.842 & 	0.688 & 	0.417 & 	0.559 &  65.036&  7.800\\
& GPT-4o-mini & 0.516& & 0.5822 & 0.9446 & 0.5874 & 0.6286 & 0.4820 & 0.4121 & 8.093 \\
& \llama 3.1-70B & 0.507 & 0.702 & 0.927 & 0.860 & 0.632 & 0.369 & 0.417 & 54.710 & 7.027  \\
& \llama 3.1-8B & 0.501 & 0.824 & 0.943 & 0.882 & 0.637 & 0.410 & 0.409 & 49.870 & 6.973 \\

\addlinespace[0.5em]

\multirow{3}{8pt}{\textbf{BookSum}} &Human& n/a & 0.344 & 	0.918 & 	0.766 & 	0.621 & 	0.349 & 	0.517 & 89.376 & 7.605 \\
& GPT-4o-mini & 0.355 & 0.385 & 0.975 & 0.831 & 0.573 & 0.441 & 0.487 & 95.169 & 7.432 \\
& \llama 3.1-70B & 0.455 & 0.544 & 0.956 & 0.842 & 0.511 & 0.314 & 0.416 & 69.068 & 6.296  \\
& \llama 3.1-8B & 0.453 & 0.634 & 0.971 & 0.842 & 0.550 & 0.394 & 0.425 & 76.472 & 6.457  \\

\addlinespace[0.5em]

\multirow{3}{8pt}{\textbf{SQuALITY}} 
& Human& n/a & 0.395 & 	0.919 & 	0.782 & 	0.565 & 	0.339 & 	0.505 & 61.257 & 5.800 \\
& GPT-4o-mini & 0.382 & 0.425 & 0.969 & 0.774 & 0.518 & 0.328 & 0.433 & 79.698 &  6.720   \\
& \llama 3.1-70B & 0.412 & 0.498 & 0.963 & 0.797 & 0.454 & 0.205 & 0.384 & 41.584 & 4.027  \\
& \llama 3.1-8B & 0.426 & 0.601 & 0.979 & 0.835 & 0.469 & 0.233 & 0.383 & 47.011 & 5.587  \\

\addlinespace[0.5em]

\multirow{3}{8pt}{\textbf{FacetSum}} 
& Human& n/a & 0.415 & 	0.940 & 	0.829 & 	0.745 & 	0.490 & 	0.584 & 72.317 &  8.147\\
& GPT-4o-mini & 0.477 & 0.483 & 0.953 & 0.685 & 0.659 & 0.526 & 0.468 & 80.937  & 7.711   \\
& \llama 3.1-70B & 0.474 & 0.622 & 0.944 & 0.899 & 0.705 & 0.507 & 0.487 & 55.197 & 6.501   \\
& \llama 3.1-8B & 0.456 & 0.565 & 0.952 & 0.894 & 0.698 & 0.471 & 0.490 & 46.081  & 6.813  \\

\addlinespace[0.5em]

\multirow{3}{8pt}{\textbf{JRC-Acquis}} 
& Human& n/a & 	0.435 & 	0.971 & 	0.860 & 	0.713 & 	0.566 & 	0.605 & 72.317 &  7.902\\
& GPT-4o-mini & 0.513 & 0.566 & 0.952 & 0.551 & 0.685 & 0.578 & 0.440 & 75.351 & 7.681  \\
& \llama 3.1-70B & 0.493 & 0.792 & 0.854 & 0.884 & 0.648 & 0.422 & 0.403 & 38.876 & 6.653  \\
& \llama 3.1-8B & 0.490 & 0.788 & 0.929 & 0.882 & 0.633 & 0.446 & 0.420 & 49.870 & 6.833 \\

\addlinespace[0.5em]

\multirow{3}{8pt}{\textbf{MultiUN}} 
& Human& n/a & 0.422 & 	0.942 & 	0.875 & 	0.605 & 	0.317 & 	0.531 & 64.540 & 8.139 \\
& GPT-4o-mini & 0.484 & 0.604 & 0.954 & 0.623 & 0.615 & 0.482 & 0.414 & 72.007 & 7.917  \\
& \llama 3.1-70B & 0.483 & 0.654 & 0.907 & 0.918 & 0.625 & 0.289 & 0.417 & 37.311 & 6.694 \\
& \llama 3.1-8B & 0.512 & 0.665 & 0.925 & 0.908 & 0.603 & 0.326 & 0.419 & 45.155 & 7.028 \\

\addlinespace[0.5em]

\multirow{3}{8pt}{\textbf{Wikipedia}} 
& Human& n/a & 0.355 & 	0.967 & 	0.817 & 	0.738 & 	0.333 & 	0.557 & 81.923 & 7.653 \\
& GPT-4o-mini & 0.471 & 0.516 & 0.960 & 0.596 & 0.687 & 0.432 & 0.432 & 76.772 & 7.609   \\
& \llama 3.1-70B & 0.467 & 0.779 & 0.877 & 0.849 & 0.638 & 0.300 & 0.377 & 55.349 & 6.493 \\
& \llama 3.1-8B & 0.476 & 0.701 & 0.940 & 0.786 & 0.606 & 0.257 & 0.378 & 51.110 & 6.014 \\

\addlinespace[0.5em]

\multirow{3}{8pt}{\textbf{GovReport}} 
& Human& n/a & 	0.397 & 	0.954 & 	0.823 & 	0.712 & 	0.405 & 	0.563 & 60.368 & 8.027 \\
& GPT-4o-mini & 0.488 & 0.521 & 0.968 & 0.605 & 0.684 & 0.521 & 0.451 & 76.887 & 7.680  \\
& \llama 3.1-70B & 0.489 & 0.638 & 0.916 & 0.872 & 0.634 & 0.425 & 0.442 & 43.421 & 6.347  \\
& \llama 3.1-8B & 0.479 & 0.531 & 0.971 & 0.881 & 0.623 & 0.384 & 0.466 & 40.544  & 7.093\\

\addlinespace[0.5em]

\multirow{3}{8pt}{\textbf{Summscreen}} 
& Human& n/a & 	0.395 & 	0.939 & 	0.739 & 	0.552 & 	0.414 & 	0.513 & 63.691& 8.373 \\
& GPT-4o-mini & 0.347 & 0.443 & 0.969 & 0.756 & 0.503 & 0.346 & 0.426 & 60.306 & 6.200  \\
& \llama 3.1-70B & 0.432 & 0.487 & 0.979 & 0.845 & 0.502 & 0.294 & 0.426 & 60.564 & 6.627  \\
& \llama 3.1-8B & 0.432 & 0.519 & 0.987 & 0.851 & 0.522 & 0.322 & 0.433 & 49.467 & 6.413  \\

\addlinespace[0.5em]

\multirow{3}{8pt}{\textbf{Seahorse}} 
& Human& n/a & 	0.336 & 	0.964 & 	0.828 & 	0.673 & 	0.533 & 	0.586 & 66.028  &  7.907\\
& GPT-4o-mini & 0.415 & 0.439 & 0.970 & 0.722 & 0.607 & 0.507 & 0.474 & 71.568 & 8.173   \\
& \llama 3.1-70B & 0.454 & 0.589 & 0.957 & 0.834 & 0.615 & 0.441 & 0.452 & 54.075 & 6.600 \\
& \llama 3.1-8B & 0.469 & 0.642 & 0.960 & 0.847 & 0.601 & 0.456 & 0.444 & 54.142 & 6.760  \\

\addlinespace[0.5em]
\bottomrule
\end{tabular}
}
\caption{Performance on the 20\% summarization task per dataset.}
\label{tab:summarization-20}
\end{table*}

\end{itemize}

\end{document}